%% file: SIGMOD.tex
\documentclass[sigconf]{acmart}
\makeatletter
\def\@ACM@checkaffil{
    \if@ACM@instpresent\else
    \ClassWarningNoLine{\@classname}{No institution present for an affiliation}%
    \fi
    \if@ACM@citypresent\else
    \ClassWarningNoLine{\@classname}{No city present for an affiliation}%
    \fi
    \if@ACM@countrypresent\else
        \ClassWarningNoLine{\@classname}{No country present for an affiliation}%
    \fi
}
\makeatother

\usepackage{booktabs}
\usepackage{url}
\usepackage{multirow}
\usepackage[shortlabels]{enumitem}
\usepackage{graphicx,subfigure}
\usepackage[shortlabels]{enumitem}
\setenumerate[1]{itemsep=0pt,partopsep=0pt,parsep=\parskip,topsep=0pt}
\setitemize[1]{itemsep=0pt,partopsep=0pt,parsep=\parskip,topsep=0pt}
\setdescription{itemsep=0pt,partopsep=0pt,parsep=\parskip,topsep=0pt}
\usepackage{caption}
\let\temp\rmdefault
\let\rmdefault\temp

\usepackage[colorinlistoftodos,prependcaption,textsize=tiny]{todonotes}
\usepackage{lipsum}                     
\usepackage{xargs}                      
 
\usepackage{colortbl} 
\usepackage{array} 
\usepackage{xcolor}

\usepackage{hyperref}
\hypersetup{
    colorlinks=true,
    linkcolor=blue,     
    urlcolor=blue
}

\newcommand{\LightCTS}{\mbox{LightCTS}}
\newcommand{\GWNet}{\textsc{GwNet}}
\newcommand{\MTGNN}{\textsc{MtGnn}}
\newcommand{\AutoCTS}{\textsc{AutoCts}}
\newcommand{\AutoCTSKDF}{\textsc{AutoCts-KDf}}
\newcommand{\AutoCTSKDP}{\textsc{AutoCts-KDp}}

\newcommand{\DCRNN}{\textsc{DcRnn}}

\newcommand{\AGCRN}{\textsc{AgCrn}}

\newcommand{\DSANet}{\textsc{DsaNet}}
\newcommand{\MAGNN}{\textsc{MaGnn}}
\newcommand{\Enhancenet}{\textsc{EnhanceNet}}
\newcommand{\FOGS}{\textsc{Fogs}}
\newcommand{\LightCTST}{\mbox{LightCTS$\backslash$T}}
\newcommand{\LightCTSM}{\mbox{LightCTS$\backslash$M}}
\newcommand{\LightCTSF}{\mbox{LightCTS$\backslash$F}}
\newcommand{\LightCTSL}{\mbox{LightCTS$\backslash$LA}}

\newcommand{\LightCTSLS}{\mbox{LightCTS$\backslash$LS}}
\newcommand{\LightCTSA}{\mbox{LightCTS\text{-}A}}
\newcommand{\LightCTSCGCN}{\mbox{LightCTS\text{-}CGCN}}
\newcommand{\LightCTSDGCN}{\mbox{LightCTS\text{-}DGCN}}

\newcommand{\Lai}[1]{{\textcolor{blue}{#1}}}

\newtheorem{finding}{\bf Observation}

\copyrightyear{2023}
\acmYear{2023} 
\setcopyright{acmcopyright}
\acmConference[SIGMOD '23]{the 2023
International Conference on Management of Data}{June 18--23, 2023}{Seattle, WA, USA}
\acmPrice{15.00}
\acmDOI{nnnnn/nnnnnn.nnnnn}

\begin{document}

\pagestyle{plain}
\pagenumbering{arabic}

\title{\LightCTS{}: A Lightweight Framework for Correlated Time Series Forecasting}

\author{Zhichen Lai$^\dag$, Dalin Zhang$^{\dag\ast}$, Huan Li$^{\ddag\ast}$, Christian S. Jensen$^\dag$, Hua Lu$^\S$, Yan Zhao$^\dag$ 
}
\affiliation{$^\dag$Department of Computer Science, Aalborg University, Denmark\\
$^\ddag$College of Computer Science and Technology, Zhejiang University, China\\
$^\S$Department of People and Technology, Roskilde University, Denmark}
\thanks{*corresponding authors: D.~Zhang (dalinz@cs.aau.dk) and H.~Li (lihuan.cs@zju.edu.cn)}

\begin{abstract}
 \input{content/0.Abstract}   
\end{abstract}


\maketitle

\input{content/1.Introduction}
\input{content/2.Preliminaries}
\input{content/3.Modeling_and_Analysis}

\input{content/4.Methodology}

\input{content/5.Experiments}

\input{content/6.Related_Work}
\input{content/7.Conclusion_and_Future_Work}

\balance

\bibliographystyle{ACM-Reference-Format}
\bibliography{references}
\end{document}

%% file: content/0.Abstract.tex
Correlated time series (CTS) forecasting plays an essential role in many practical applications, such as traffic management and server load control. Many deep learning models have been proposed to improve the accuracy of CTS forecasting. However, while models have become increasingly complex and computationally intensive, they struggle to improve accuracy. Pursuing a different direction, this study aims instead to enable much more efficient, lightweight models that preserve accuracy while being able to be deployed on resource-constrained devices. To achieve this goal, we characterize popular CTS forecasting models and yield two observations that indicate directions for lightweight CTS forecasting. On this basis, we propose the \LightCTS{} framework that adopts {plain stacking} of temporal and spatial operators instead of alternate stacking which is much more computationally expensive. Moreover, \LightCTS{} features light temporal and spatial operator modules, called L-TCN and GL-Former, that offer improved computational efficiency without compromising their feature extraction capabilities. \LightCTS{} also encompasses a {last-shot compression} scheme to reduce redundant temporal features and speed up subsequent computations. Experiments with single-step and multi-step forecasting benchmark datasets show that \LightCTS{} is capable of nearly state-of-the-art accuracy at much reduced computational and storage overheads.

%% file: content/1.Introduction.tex
\section{Introduction}
\label{sec:intro}

\begin{figure}[]
        \includegraphics[width=.8\linewidth]{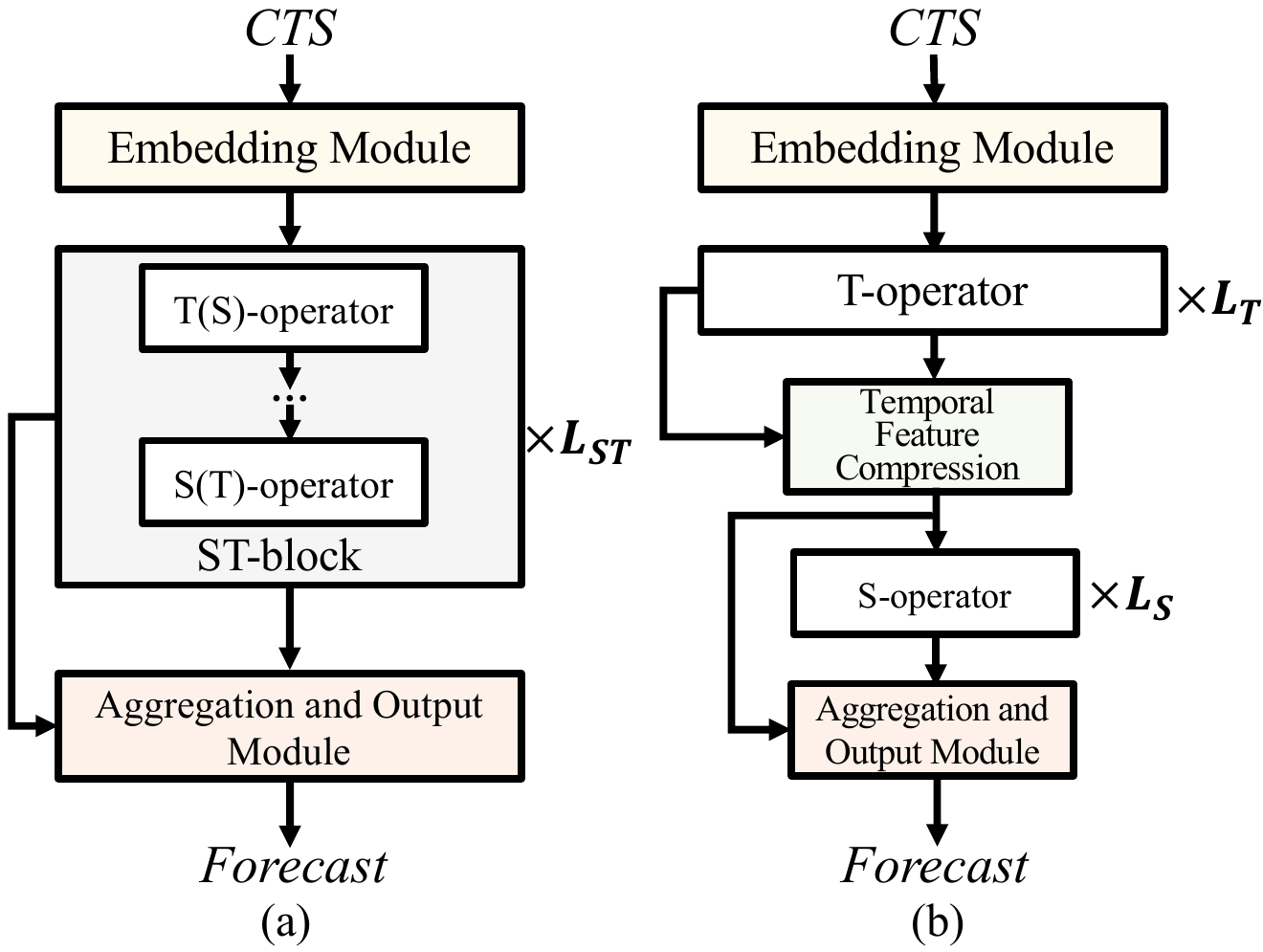}
	\caption{DL-based CTS forecasting frameworks using (a) alternate stacking and (b) plain stacking.}
	\label{fig:ctsoverview}
\end{figure}

Driven in part by the availability of increasingly advanced and affordable sensor technologies, cyber-physical systems (CPSs)~\cite{derler2011modeling} are being deployed at a rapid pace. In a typical CPS, multiple sensors sample physical processes of interest and emit multiple time series with correlations. One example is sensors that sample power production by photovoltaic installations in a geographical region~\cite{lai2018modeling}. 
Extracting and exploiting correlations in correlated time series (CTS) is important in many applications, such as the forecasting of traffic situations~\cite{cirstea2021enhancenet,wu2021autocts,papadimitriou2006optimal,yuan2020effective}, air quality~\cite{du2019deep}, server loads~\cite{faloutsos2019classical,ma2018query}, social activity~\cite{zhu2002statstream}, and wind farm maintenance~\cite{cheng2021novel}. In this study, we focus on CTS forecasting. 
One significant application occurs in the predictive maintenance of wind turbines~\cite{cheng2021novel}, which are often deployed in remote and harsh locations. Accurate and instant forecasts of a turbine's operating status, e.g., covering pitch speed and active power, can enable identification of potential failures, thereby enabling timely maintenance, and thus reducing regular maintenance costs, decreases in generated power, and potential safety hazards~\cite{lai2022multiscale}. Hence, forecasting has attracted extensive research attention.

Deep learning (DL) techniques have recently shown impressive CTS forecasting performance. A variety of DL modules, such as convolutional neural networks (CNNs)~\cite{yu2018spatio,wu2019graph,wu2020connecting,guo2019attention,huang2020lsgcn,chen2022multi,rao2022fogs}, recurrent neural networks (RNNs)~\cite{bai2020adaptive,chen2020multi,li2018diffusion,chang2018memory,shih2019temporal}, graph convolutional networks (GCNs)~\cite{li2018diffusion, pan2021autostg,wu2019graph,wu2021autocts,guo2019attention,chen2022multi,rao2022fogs}, 
and Transformers~\cite{xu2020spatial,park2020st,zhou2021informer,wu2021autocts}, are used to construct operators for extracting temporal features from individual time series or spatial features across correlated time series. 
These two categories of operators are referred to as temporal operators (\textbf{T-operators}) and spatial operators (\textbf{S-operators}) (see the categorization in Table~\ref{tab:component}). Studies~\cite{yu2018spatio,wu2021autocts} show that such operators are effective at feature extraction and enable state-of-the-art CTS forecasting accuracy. We analyze the commonalities of these DL-based models and present a generic framework as shown in Figure~\ref{fig:ctsoverview}(a).
The framework starts with an embedding module that ingests the raw CTS data; then, multiple spatio-temporal blocks (ST-blocks) are stacked, each consisting of a sequence of alternate S- and T-operators for extracting high-order spatio-temporal features (ST-features);
the framework ends with a module that aggregates ST-features at different depths with residual connections and outputs a forecast.

However, DL-based CTS models are often large, and inferencing is often computationally expensive. This limits the possibilities of deploying CTS forecasting on resource-constrained edge computing devices, which is otherwise attractive in CPS applications due to the decentralized computation and potentially increased service responsiveness~\cite{shi2016edge}. 
As an example, there is a pressing need to monitor and forecast the working status of wind turbines in real time for maintenance~\cite{cheng2021novel}. However, wind turbines are often deployed offshore or in high latitudes, and transmitting their operational data to a remote data center is costly, lagging, and fragile to the quality of networking.
To detect potential faults in a timely manner and respond on board, an appealing solution is to place a lightweight model on an edge device operating locally.  Microcontroller units (MCUs) are widely used edge devices in industry due to their stability and low cost~\cite{sudharsan2021ml}. However, they have very limited memory. The popular STM32F4 series of MCUs have up to 3Mb of memory, which is insufficient for the deployment of state-of-art CTS models like \GWNet{} \cite{wu2019graph}. Moreover, as existing CTS models are not specifically designed for lightweight applications, simply lightening these models degrades their performance dramatically (see Table \ref{tab:memory}).

Moreover, we observe that although recent studies on CTS forecasting focus mainly on improving accuracy, progress has almost come to a standstill. For example, \AutoCTS{}~\cite{wu2021autocts}, a state-of-the-art forecasting model, improves \GWNet{}~\cite{wu2019graph}, a previous state-of-the-art model, by up to 0.06 miles per hour (mph) in terms of mean absolute error (MAE) on traffic speed forecasting. \AutoCTS{} models are much larger and are obtained through neural architecture search, which involves the training and evaluation of thousands of large models, which may take up to hundreds of GPU hours, incurring considerable $CO_2$ emissions~\cite{chen2021bn}. This situation, characterized by increasingly larger and computationally expensive models with diminishing accuracy improvements, motivates a different direction, where we instead aim to achieve lightweight models with competitive accuracy. This will enable edge computing as well as overall emissions savings~\cite{shi2016edge}.

Although several lightweight techniques have been proposed in computer vision~\cite{tan2019efficientnet,zhang2018shufflenet,han2020ghostnet,li2021dynamic}, these techniques are not readily applicable to CTS forecasting. A key reason is that lightweight computer vision models focus mainly on simplifying 2D and 3D convolutions for image and video data, while CTS models involve instead 1D convolution of temporal features and graph convolution of spatial features.
In addition, recent lightweight Transformers~\cite{mehta2021mobilevit,liu2021swin} reduce the computational cost of the self-attention mechanism by utilizing the similarities among adjacent and multi-scale image patches, which are not applicable to CTS data.

We propose \LightCTS{}, a framework that enables lightweight CTS forecasting at significantly reduced computational cost while retaining forecasting accuracy comparable to the state-of-the-art. 
We start with a comprehensive analysis of existing CTS models, placing them in a generic framework (see Figure~\ref{fig:ctsoverview}(a)) and scrutinizing the computational and storage overheads of their components, both theoretically and empirically.
The analysis yields important observations (see Section~\ref{sec:modelandanalysis}) and points to two directions for achieving lightness, namely 1) simplifying computations associated with ST-feature extraction and 2) optimizing the generic CTS architecture and  compressing redundant temporal dimensions for costly S-operators.

By following these directions, \LightCTS{} offers a set of novel lightweight techniques.
First, \LightCTS{} includes a novel T-operator module called \emph{Light Temporal Convolutional Network} (L-TCN) and a novel S-operator module called \emph{GlobalLocal TransFormer} (GL-Former) for temporal and spatial feature extraction, respectively. Both L-TCN and GL-Former adopt grouping strategies to reduce the full connections between adjacent layers to local in-group connections, thus achieving lower complexity than the vanilla TCN and Transformer.
Moreover, \LightCTS{} adopts a simple but effective \emph{plain stacking} based architecture (see Figure~\ref{fig:ctsoverview}(b)) that decouples temporal and spatial feature extraction and renders the compression and reduction of intermediate features more flexible. Along with plain stacking, a \emph{last-shot compression} scheme is employed that retains only the last time-step slice of temporal features extracted by T-operators.
This scheme reduces the features that are fed to subsequent components with only minor information loss, as TCNs tend to capture the most significant features in the last time step~\cite{lea2016temporal}.
The plain stacking and last-shot compression combine to considerably lower the computational and storage overheads of the subsequent S-operators as well as of the aggregation and output module.

Considering both single-step and multi-step CTS forecasting, we conduct extensive experiments to evaluate \LightCTS{} on six benchmark datasets. 
We find that \LightCTS{} achieves accuracy comparable to those of state-of-the-art models, but with much lower computational and storage overheads.
We have made our implementation publicly available\footnote{\url{https://github.com/AI4CTS/lightcts}}.

The contributions of the paper are summarized as follows.
\begin{itemize}[leftmargin=*]

    \item We propose \LightCTS{}, a novel lightweight CTS forecasting framework. To the best of our knowledge, this is the first study of lightweight DL-based CTS forecasting.
    
    \item We analyze the architectures, S/T-operators, and resource costs of mainstream CTS models, and identify key opportunities for achieving lightness.

    \item We contribute L-TCN and GL-Former, two novel lightweight T- and S-operator modules, targeting the extraction of ST-features of CTS. We also propose a plain stacking pattern and a last-shot compression scheme, targeting a reduction of the sizes of the inputs to S-operators and the aggregation and output module.

    \item We report on experimental findings for different tasks, offering evidence that \LightCTS{} is capable of state-of-the-art accuracy while reducing computational costs and model sizes very substantially.
 
\end{itemize}

Section~\ref{sec:preliminaries} covers the definition of CTS and its forecasting tasks; Section~\ref{sec:modelandanalysis} analyzes the commonalities of existing CTS models;
Section~\ref{sec:lightCTS} details the design of KDCTS;
Section~\ref{sec:experiments} reports on the experimental study;
Section~\ref{sec:related} covers related work on CTS forecasting and knowledge distillation of DL models; finally, Section~\ref{sec:conclusion} concludes and presents research directions.

%% file: content/2.Preliminaries.tex
\section{Preliminaries}
\label{sec:preliminaries}

This section covers preliminaries of CTS and formalizes the problem of CTS forecasting.
Frequently used notations are summarized in Table~\ref{tab:notation}.

\begin{table}
\caption{Summary of notation.}
\label{tab:notation}
\centering
\resizebox{\linewidth}{!}{

\begin{tabular}{|c|c|}
\hline
\textbf{Notation} & \textbf{Description}                                 \\ \hline
$X$                 & An indexed set of correlated time series (CTS) \\
$\mathtt{N}$        & Number of time series in $X$   \\
$\mathtt{T}$        & Number of time steps in $X$ \\
$\mathtt{D}$        & Embedding size of S/T-operators  \\
$\mathtt{P}$        & Number of historical time steps used in CTS forecasting \\
$\mathtt{Q}$        & Number of future time steps of CTS forecasting \\
\hline
\end{tabular}
}
\end{table}

\subsection{Correlated Time Series}
\label{ssec:cts_definition}

In a cyber-physical system~\cite{derler2011modeling}, $\mathtt{N}$ devices each generate timestamped data, yielding $\mathtt{N}$ time series. The time series are called \textbf{correlated time series} (CTS)~\cite{wu2021autocts}, denoted as $X \in \mathbb{R}^{\mathtt{N} \times \mathtt{T} \times \mathtt{F}}$, where $\mathtt{T}$ and $\mathtt{F}$ denote the number of time steps and the number of sensor measurements per time step, respectively. For example, in a wind turbine farm consisting of 30 turbines, each turbine may emit wind speed and wind direction measurements at each time step; thus, if the system emits measurements for 500 time steps, we get $X \in \mathbb{R}^{ 30 \times 500 \times 2}$.

Given the $\mathtt{N}$ time series, two kinds of correlations occur: temporal correlations within time series and spatial correlations across different time series.
On the one hand, consecutive measurements in a time series are naturally correlated.
On the other hand, concurrent measurements by different devices may be correlated due to, e.g., the spatial proximity of the devices. For example, traffic flows reported by sensors on connected road segments are naturally correlated~\cite{pedersen2020anytime,wu2021autocts}.

\subsection{CTS Forecasting Problems}
\label{ssec:cts_forecasting}

We consider single-step and multi-step CTS forecasting. 
First, \textbf{single-step CTS forecasting} aims to predict the $\mathtt{Q}$-th future time step ($\mathtt{Q} \geq 1$); formally,
\begin{equation}
\hat{X}_{t+\mathtt{P+Q}} \gets \mathcal{SF}(X_{t+\mathtt{1}}, \ldots, X_{t+\mathtt{P}}),
\end{equation}
where $t$ indexes the beginning time step, $\mathtt{P}$ is the number of historical time steps used for forecasting, $\hat{X}_{t+\mathtt{P+Q}}$ denotes the predicted CTS at the future $(t+\mathtt{P+Q})$-th time step, and $\mathcal{SF}$ denotes a single-step CTS forecasting model.

Next, \textbf{multi-step CTS forecasting} aims to predict $\mathtt{Q}$ ($\mathtt{Q} > 1$) consecutive future time steps in one pass; formally,
\begin{equation}
\{\hat{X}_{t+\mathtt{P+1}}, \ldots, \hat{X}_{t+\mathtt{P+Q}} \} \gets \mathcal{MF}(X_{t+\mathtt{1}}, \ldots, X_{t+\mathtt{P}}),
\end{equation}
where $\mathcal{MF}$ denotes a multi-step CTS forecasting model.

For both problems, it is essential to extract the temporal dynamics in each time series and the spatial correlations among different time series from the historical data. To this end, deep learning (DL) techniques with powerful temporal and spatial feature extraction capabilities have been used widely in CTS models.
A review of existing DL-based CTS models is provided in Section~\ref{sec:related}. Due to the characteristics of the neural network operators used to extract spatial and temporal features, the training and inferencing of DL-based CTS models incur considerable computational and storage overheads. In this study, \emph{we aim to enable lightweight CTS forecasting models (i.e., models with fewer computations and parameters) with forecasting accuracy comparable to the state-of-the-art CTS forecasting models.}

%% file: content/3.Modeling_and_Analysis.tex
\section{Analyses and Directions}
\label{sec:modelandanalysis}

In Section~\ref{ssec:analysis}, we place existing DL-based CTS modeling proposals in a generic framework, map the complexity of their internal operators, and investigate the computational and storage overheads of representative models.
Based on this analysis, we identify directions for achieving a lightweight CTS framework in Section~\ref{ssec:directions}.

\subsection{Analysis of Existing CTS Models}
\label{ssec:analysis}

\subsubsection{Generic Framework}\label{sssec:framework}
To gain insight into the prospects of lightening DL-based CTS models, we consider representative proposals~\cite{li2018diffusion,yu2018spatio,wu2019graph,bai2020adaptive,wu2020connecting,xu2020spatial,park2020st,guo2019attention,chen2020multi,shih2019temporal,wu2021autocts,zhou2021informer,grigsby2021longrange,chang2018memory,chen2022multi,huang2020lsgcn,cirstea2021enhancenet,rao2022fogs}.
Figure~\ref{fig:ctsoverview}(a) shows a generic framework for these models.
Generally, a CTS model has three components: 
(1) an \emph{embedding module} that transforms the raw CTS into latent representations; 
(2) a stack of \emph{spatio-temporal blocks} (ST-blocks) that extract spatial and temporal correlations as high-order features; and (3) an \emph{aggregation and output module} that aggregates ST-features from the ST-blocks and outputs the result, which is either a single-step or a multi-step forecast (see Section~\ref{ssec:cts_forecasting}).

Being responsible for extracting temporal and spatial correlations, ST-blocks make up the key component of a CTS model.
Typically, an ST-block includes alternating stacks of \emph{temporal operators} (T-operators) and \emph{spatial operators} (S-operators).
For example, the alternate stacking pattern can be $ \langle T, S \rangle$, $ \langle S, T \rangle$, $ \langle T, S, T \rangle$, etc. The S/T-operators are the basic ingredients for extracting comprehensible features.

\subsubsection{S/T-operators}
We proceed to study the S/T-operators employed by state-of-the-art models~\cite{li2018diffusion,yu2018spatio,wu2019graph,bai2020adaptive,wu2020connecting,xu2020spatial,park2020st,guo2019attention,chen2020multi,shih2019temporal,wu2021autocts,zhou2021informer,grigsby2021longrange,chang2018memory,chen2022multi,huang2020lsgcn,cirstea2021enhancenet,rao2022fogs}.
Specifically, we analyze each operator's time complexity in terms of \emph{FLOPs} (floating-point operations) and space complexity in terms of \emph{the number of model parameters}.  
Referring to Table~\ref{tab:component}, we categorize popular S- and T-operators into different families based on the base operators that they extend.
Considering that there are only minor differences between the operators in a family (e.g., applying different attention mechanisms or convolution kernels), we report on the time and space complexity of the base operators in Table~\ref{tab:component}.
We refer interested readers to the supplemental material~\cite{lightcts} for a detailed analysis of base operators.

\begin{table}[]
\setlength\tabcolsep{1em}
\caption{Categorization and analysis of ST-block operators.}\label{tab:component}
\resizebox{\linewidth}{!}{
\begin{tabular}{c|cc|cc}
\toprule
Type & Family    & Literature & Time Complexity                           & Space Complexity     \\ \midrule
\multicolumn{1}{c|}{\multirow{6}{*}{\rotatebox{90}{T-operator}}} &
  CNN &
  \multicolumn{1}{c|}{\begin{tabular}[c]{@{}c@{}}~\cite{yu2018spatio,wu2019graph,guo2019attention,wu2020connecting,cirstea2021enhancenet}\\ ~\cite{huang2020lsgcn,wu2021autocts,chen2022multi,rao2022fogs}\end{tabular}}   &
  $\mathcal{O}( \mathtt{D}^2 \cdot \mathtt{N} \cdot \mathtt{P})$ &
  $\mathcal{O}(\mathtt{D}^2 )$  \\ \cmidrule{2-5} 
  &
  RNN &
  \multicolumn{1}{c|}{\begin{tabular}[c]{@{}c@{}}~\cite{bai2020adaptive,chen2020multi,chang2018memory}\\ ~\cite{li2018diffusion,cirstea2021enhancenet,shih2019temporal}\end{tabular}}   &
  $\mathcal{O}(\mathtt{D}^2 \cdot \mathtt{N} \cdot \mathtt{P})$ &
  $\mathcal{O} (\mathtt{D}^2)$ \\ \cmidrule{2-5} 
  &
  Transformer &
  \multicolumn{1}{c|}{\begin{tabular}[c]{@{}c@{}}~\cite{xu2020spatial,park2020st}\\ ~\cite{zhou2021informer,wu2021autocts}\end{tabular}} &
  $\mathcal{O}( \mathtt{D} \cdot \mathtt{N} \cdot \mathtt{P} \cdot (\mathtt{P} + \mathtt{D}) )$ & 
  $\mathcal{O}( \mathtt{D}^2) $                 \\ \midrule
\multicolumn{1}{c|}{\multirow{4}{*}{\rotatebox{90}{S-operator}}} &
  GCN &
  \multicolumn{1}{c|}{\begin{tabular}[c]{@{}c@{}}~\cite{li2018diffusion, guo2019attention,wu2019graph,wu2021autocts} \\ ~\cite{cirstea2021enhancenet,chen2022multi,rao2022fogs}\end{tabular}}   &
  $\mathcal{O}(\mathtt{D} \cdot \mathtt{N} \cdot \mathtt{P} \cdot (\mathtt{N} + \mathtt{D}))$ & $ \mathcal{O}(\mathtt{D}^2)$ \\ \cmidrule{2-5}
  &
  Transformer 
  &
  \multicolumn{1}{c|}{\begin{tabular}[c]{@{}c@{}}~\cite{xu2020spatial,park2020st}\\ ~\cite{grigsby2021longrange,wu2021autocts}\end{tabular}}   &
  $\mathcal{O}(\mathtt{D} \cdot \mathtt{N} \cdot \mathtt{P} \cdot (\mathtt{N} + \mathtt{D}))$ &
  $\mathcal{O}(\mathtt{D}^2)$ \\ \bottomrule
\end{tabular}
}
\vspace*{0.1em}
\small{\hfill Embedding size $\mathtt{D}$; time series number $\mathtt{N}$; historical time steps $\mathtt{P}$.}
\end{table}

Three main \textbf{T-operator} families are identified: 
(1) \emph{CNN-based T-operators} \cite{yu2018spatio,wu2019graph,wu2020connecting,guo2019attention,huang2020lsgcn,chen2022multi,cirstea2021enhancenet,rao2022fogs}, specifically Temporal Convolutional Networks (TCNs), that apply dilated causal convolutions to time series data;
(2) \emph{RNN-based T-operators}, such as long short term memory networks (LSTMs)~\cite{shih2019temporal} and gated recurrent unit networks (GRUs)~\cite{bai2020adaptive,chen2020multi,chang2018memory,li2018diffusion,cirstea2021enhancenet}, that process time series based on a recursive mechanism;
and (3) \emph{Transformer-based T-operators}~\cite{xu2020spatial,park2020st,zhou2021informer,wu2021autocts} that adopt the attention mechanism to establish self-interactions of input time steps, enabling weighted temporal information extraction over long sequences.
While all T-operator families have the same space complexity, the time complexity of the operators in the Transformer family is larger than those of the operators in the CNN and RNN families because of their large-size matrix multiplication~\cite{wu2021autocts}. Further, although the RNN family operators have the same time complexity regarding FLOPs as the CNN family operators, the former adopt a sequential computation scheme that significantly reduces the actual efficiency. 
Thus, CNN-based T-operators are the most promising for lightweight CTS models. Recent TCN models, including \GWNet{}~\cite{wu2019graph}, \MTGNN{}~\cite{wu2020connecting}, and \FOGS{}~\cite{rao2022fogs}, achieve the state-of-the-art accuracies.

There are roughly two \textbf{S-operator} families: 
(1) \emph{GCN-based S-operators}, specifically Chebyshev GCNs~\cite{guo2019attention,cirstea2021enhancenet,chen2022multi,rao2022fogs} or Diffusion GCNs~\cite{li2018diffusion,wu2019graph,wu2021autocts}, utilize predefined or learned spatial adjacency matrices to capture high-order spatial correlations and
(2) \emph{Transformer-based S-operators}~\cite{xu2020spatial,park2020st,grigsby2021longrange,wu2021autocts} cast attention operations across different time series to obtain their weighted spatial correlations.
Theoretically, GCNs and Transformers incur the same space and time complexities for S-operators (see Table~\ref{tab:component}). Moreover, no existing studies compare their CTS forecasting performance in the same setting. We thus include experiments that compare these two S-operators in our \LightCTS{} framework. The results, in Section \ref{ssec:ablation} and Table \ref{tab:ablation}, show that a Transformer-based S-operator achieves the best accuracy in our framework.

\subsubsection{FLOP and Parameter Use in CTS Models}\label{sssec:distributions}
To investigate the resource consumption of each component of CTS models, we analyze FLOPs and parameters via a case study. 
We select three representatives, namely \FOGS{}~\cite{rao2022fogs} as the most accurate model, \MTGNN{}~\cite{wu2020connecting} as the most lightweight model, and \GWNet{}~\cite{wu2019graph} as a widely used model.
They are also all included for comparison in the experimental study in Section~\ref{sec:experiments}.
We select the METR-LA dataset (see Section~\ref{sssec:multi_step_datasets}) and use the architectures reported by the original papers. Table~\ref{tab:distributions} shows the distribution of FLOPs and parameters associated with different components of CTS models, namely the embedding module, the T-operators and S-operators in the ST-component, and the aggregation and output module.

\begin{table}[!htbp]
\caption{Distribution of FLOPs and parameters in CTS models.}
\label{tab:distributions}
\centering
\footnotesize

\resizebox{\linewidth}{!}{
\begin{tabular}{cc|cccc}
\toprule
\multirow{2}{*}{Model} & \multirow{2}{*}{Metric} & Input & T- & S- & Aggregation \\
                   &            &  Embedding       & operators & operators & and Output \\ \midrule
\multirow{2}{*}{\FOGS{}} & FLOPs      & 0.01\% & 2.13\%   & 95.72\%   & 2.15\%    \\
         & Parameters & 0.01\% & 3.25\%   & 77.19\%   & 19.55\%    \\ \midrule
 \multirow{2}{*}{\MTGNN{}} & FLOPs& 0.12\% & 23.54\%    & 70.19\%    & 6.15\%    \\
                   & Parameters & 0.02\% & 75.52\%   & 9.49\%   & 14.97\%     \\ \midrule
 \multirow{2}{*}{\GWNet{}} & FLOPs      & 0.03\% & 6.70\%   & 75.23\%   & 18.04\%    \\
         & Parameters & 0.03\% & 10.77\%   & 19.94\%   & 69.26\% \\ \bottomrule
\end{tabular}
}
\end{table}
As expected, a significant portion of FLOPs and parameters occur in the S- and T-operators that make up the core component of a CTS model. 
Moreover, S-operators consume many more FLOPs than T-operators. 
Surprisingly, the aggregation and output module is also responsible for many FLOPs and parameters.
This is likely because the aggregation and output module has to process massive ST-features extracted by all ST-blocks. 

\subsection{Observations and Resulting Directions}
\label{ssec:directions}

We highlight the main observations from the above analyses and identify promising directions for designing \LightCTS{}.

\begin{finding}\label{finding1}
S- and T-operators, which make up the main component of CTS models (Section \ref{sssec:framework}), incur significant computational and storage overheads (Table \ref{tab:distributions}). Their time and space complexities are both proportional to $\mathtt{D}^2$ (Table \ref{tab:component}).
\end{finding}

Observation~\ref{finding1} indicates that it is essential to lighten S- and T-operators.
As presented in Table \ref{tab:component}, the time and space complexities of all S/T-operators are positively correlated with $\mathtt{N}$, $\mathtt{P}$, and $\mathtt{D^2}$. The numbers of time series $\mathtt{N}$ and historical time steps $\mathtt{P}$ are decided by the raw CTS data and should be left unaltered in a CTS model.
Therefore, manipulating the embedding size $\mathtt{D}$ of S/T-operators is a direction for achieving lightness. Many studies~\cite{wu2021autocts,chen2020multi,wu2020connecting}, however, have shown that simply reducing $\mathtt{D}$ inevitably degrades forecasting accuracy.
Instead, we propose to simplify and reduce the neural network computations associated with $\mathtt{D}$, to be detailed in Sections~\ref{ssec:ltcn} and~\ref{ssec:ltransformer} for T- and S-operators, respectively.

\begin{finding}\label{finding2}
S-operators consume many more resources, especially FLOPs, than T-operators (Table \ref{tab:distributions}).
\end{finding}

One reason for Observation ~\ref{finding2} is that S-operators usually have higher time complexity than T-operators (see Table \ref{tab:component}). The complexity of S-operators can be reduced by manipulating $\mathtt{D}$ as discussed for Observation \ref{finding1}. Another major reason is that the input to S-operators must retain the temporal dimensions even if these contribute little to extracting spatial correlations. This is a consequence of the alternate stacking pattern of the existing generic CTS framework (see Section \ref{sssec:framework}). 
Specifically, since executions of S- and T-operators intermixed, some T-operators occur after S-operators. To allow such T-operators to properly extract temporal features, the temporal dimensions must be preserved.
Although T-operators also receive redundant spatial information due to intermixing, S-operators are affected more due to their higher time complexity. 
Moreover, the spatial dimension $\mathtt{N}$ cannot be compressed as the aim is to forecast future values for all $\mathtt{N}$ time series. 

This observation suggests that a new stacking pattern of S/T-operators that compresses temporal features before applying S-operators. 
We thus propose to design a new stacking pattern that decouples T-operators and S-operators, to be detailed in Section \ref{ssec:plainstack}.
In addition, we devise a temporal feature compression scheme to condense the input feature maps for S-operators while preserving key temporal information to be used in the final forecasting, to be detailed in Section \ref{ssec:last_shot}.
As mentioned in Section~\ref{sssec:distributions}, the aggregation and output module takes all features extracted by ST-blocks as input, which incurs considerable FLOPs and parameters.
As a by-product of the temporal feature compression, the intermediate T- and S-features can be downsized, which leads to a significant reduction in FLOPs and parameters for the aggregation and output module.

%% file: content/4.Methodology.tex
\section{Construction of \LightCTS{}}
\label{sec:lightCTS}

We start by presenting the new \emph{plain stacking} pattern in Section~\ref{ssec:plainstack}; next, we detail the light T-operator module in Section~\ref{ssec:ltcn} and its subsequent \emph{last-shot compression} in Section~\ref{ssec:last_shot}; we then present the light S-operator module in Section~\ref{ssec:ltransformer} and the assembly of \LightCTS{} in Section~\ref{ssec:assemble}.

\subsection{Architecture with Plain Stacking}
\label{ssec:plainstack}

As illustrated in Figure~\ref{fig:ctsoverview}(a), the conventional CTS architecture relies on ST-blocks, each of which stacks S- and T-operators alternately.
Such an alternate stacking pattern maintains a feature size of $\mathtt{N} \times \mathtt{P} \times \mathtt{D}$ throughout feature extraction.
Indeed, the temporal dimension $\mathtt{P}$ is only considered in T-operators and is disregarded in S-operators.
In other words, the output representation unnecessarily increases the computational and storage overheads of S-operators. 

We deviate from the alternate stacking pattern and instead combine a T-operator module consisting of $\mathtt{L}_T$ T-operators and an S-operator module consisting of $\mathtt{L}_S$ S-operators serially, yielding what we call the \emph{plain stacking} scheme.
Figure~\ref{fig:ctsoverview}(b) presents the novel architecture, which enjoys several benefits.
First, temporal and spatial feature extraction are decoupled, allowing compression of the temporal dimension before applying the more complex S-operators (see Observation~\ref{finding2}).
In particular, we propose a \emph{last-shot compression} scheme (to be detailed in Section~\ref{ssec:last_shot}) that reduces the output size of temporal feature extraction from $\mathtt{N} \times \mathtt{P} \times \mathtt{D}$ to $\mathtt{N} \times \mathtt{D}$.

Moreover, the computational overhead of the aggregation and output module is also reduced by the feature compression.
Indeed, the architecture further reduces costs by taking as input only the final features of the temporal and spatial feature extraction phases, instead of the features from all stacked ST-blocks in the conventional architecture.

The plain stacking architecture does not lower the effectiveness of feature extraction but reduces computational costs, as will be shown in Section~\ref{ssec:ablation}.
To construct a specific \LightCTS{} model using the new architecture, we design a light T-operator module L-TCN and a light S-operator module GL-Former, presented in Sections~\ref{ssec:ltcn} and~\ref{ssec:ltransformer}.

\subsection{L-TCN}
\label{ssec:ltcn}

\subsubsection{Background of TCN}
\label{ssec:bgtcn}
We propose an {L}ight {T}emporal {C}onvolutional {N}etwork (L-TCN) that is based on the Gated TCN~\cite{wu2019graph}, which incorporates the gating mechanism into standard TCNs to control the temporal information flow. 
A TCN adopts dilated causal convolutions (DCC)~\cite{DBLP:journals/corr/YuK15,DBLP:conf/ssw/OordDZSVGKSK16} to capture both long- and short-term temporal patterns in a non-recursive fashion, thus alleviating the issue of gradient explosion in RNNs. 

\begin{figure}
	\begin{center}
		\includegraphics[width=1\linewidth]{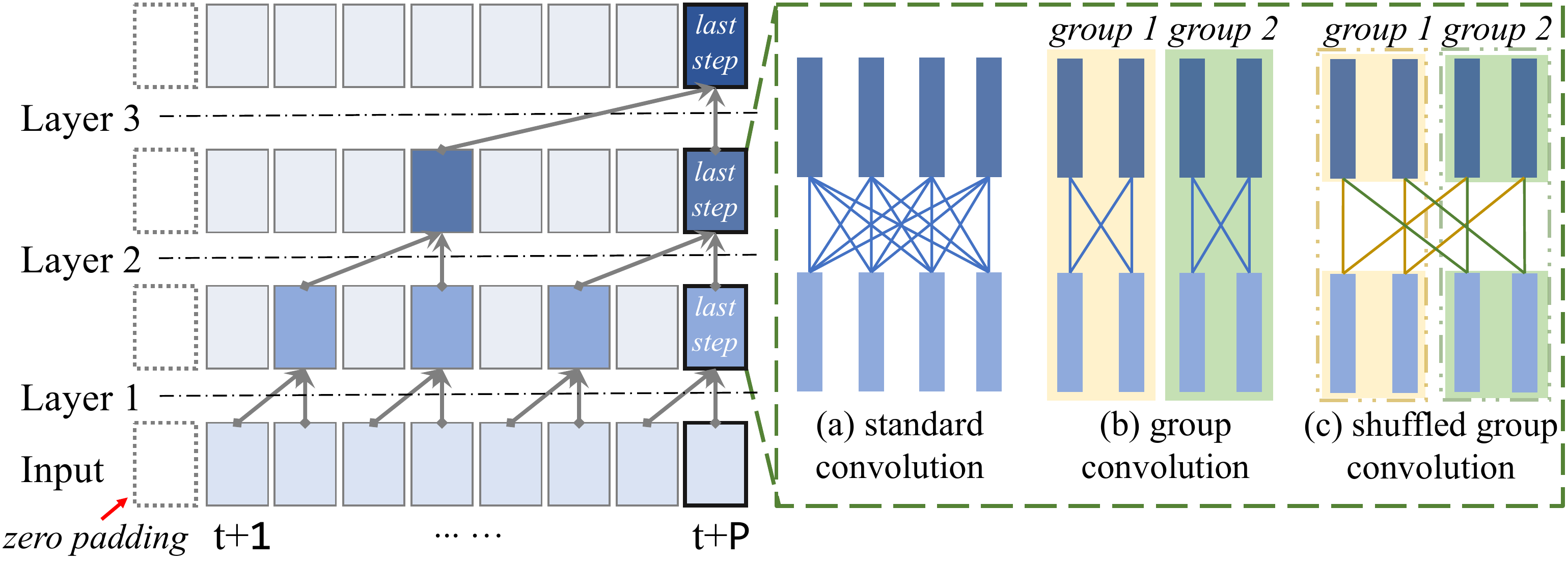}
	\end{center}
	\caption{A TCN with layers of (a) standard convolution (TCN), (b) group convolution (GTCN), and (c) shuffled group convolution (SGTCN). The TCN consists of three layers with dilation rates $\delta = \{1, 2, 4\}$ and the kernel size $\mathtt{K}$ = 2.}
	\label{fig:tcn}
\vspace{-0.5cm}
\end{figure}

Figure~\ref{fig:tcn} (left) illustrates the basic structure of a standard TCN with three DCC layers. 
In Layer 1, a convolutional filter slides over the input without skipping any values in the filter (i.e., \emph{dilation rate} $\delta$ = 1); in Layer 2, a convolutional filter that skips one value in the middle (i.e., $\delta$ = 2) is applied to the output of Layer 1; in Layer 3, convolutional operations are performed with three skipped values (i.e., $\delta$ = 4) in the filter. 
By stacking multiple DCC layers, a.k.a. TCN layers, with gradually increased dilation rates, a TCN T-operator module with exponentially enlarged receptive fields is built.
As seen in Figure~\ref{fig:tcn} (left), the receptive field of the last layer's rightmost node can cover the entire time series length of the input data, which implies that the plain stacking does not ignore important spatio-temporal correlations.

With the raw input CTS ${X} \in \mathbb{R}^{\mathtt{N} \times \mathtt{P} \times \mathtt{F}}$ being embedded into the latent representation ${H} \in \mathbb{R}^{\mathtt{N} \times \mathtt{P} \times \mathtt{D}}$ by the embedding module, a TCN layer cast on ${H}$ is formalized as follows.

\begin{equation*}
\begin{aligned}
    & \operatorname{TCN}(H \mid \delta, \mathtt{K}) = H',\;\text{where} \\
    & H'[i;t;d] = \sum\nolimits_{k=0}^{\mathtt{K}-1} \big( H[i;{t- \delta \times k};:] \cdot \text{W}^d[k;:] \big)
\end{aligned}
\end{equation*}

is the $d$-th ($d \in [0,\mathtt{D})$) output feature map at time step $t$ ($t \in [0,\mathtt{P})$) of the $i$-th ($i \in [0,\mathtt{N})$) time series, $\text{W}^d \in \mathbb{R}^{\mathtt{D} \times \mathtt{K}}$ is the $d$-th convolutional filter, and $\mathtt{K}$ (often as small as 2 or 3) and $\delta$ are the kernel size and dilation rate, respectively.
To keep the temporal dimension of $\mathtt{P}$ constant in the output, zero padding is applied to the input of each layer~\cite{DBLP:conf/ssw/OordDZSVGKSK16}.

\subsubsection{Lightening TCN}
\label{ssec:lighteningtcn}
Previous studies~\cite{wu2020connecting,chen2022multi} show that directly reducing the embedding size $\mathtt{D}$ inevitably lowers the representation capability of the model, resulting in subpar accuracy. 
Therefore, we propose instead to lighten the standard TCN using a {\it grouping} strategy. This is motivated by observations of previous studies~\cite{xie2017aggregated,zhang2018shufflenet} that a TCN has redundant connections between adjacent TCN layers and thus can be optimized.
Specifically, the standard TCN layer in Figure~\ref{fig:tcn}(a) has full connections between input and output feature maps, while the grouping strategy in Figure~\ref{fig:tcn}(b) first partitions the input TCN feature maps into $\mathtt{G}^T$ equal-sized, consecutive, and non-overlapping groups, and then performs convolutions within the groups.
A group TCN layer, i.e., a TCN layer with the grouping strategy, is represented as follows.
\begin{equation*}
    \operatorname{GTCN}(H \mid \mathtt{G}^T) = \operatorname{concat}( \{ \operatorname{TCN}(H_j\mid \delta, \mathtt{K})\}_{j = 1}^ {\mathtt{G}^T}),    
\end{equation*}
where $H_j=[:,:,\frac{\mathtt{D}\times(j-1)}{\mathtt{G}^T}:\frac{\mathtt{D} \times j}{\mathtt{G}^T}]\in\mathbb{R}^{\mathtt{N}\times\mathtt{P}\times\frac{\mathtt{D}}{{\mathtt{G}^T}}}$ is the $j$-th group of input feature maps and $\operatorname{concat}(\cdot)$ denotes the concatenation operation. The time and space complexity of each group are $\mathcal{O}\big( (\frac{\mathtt{D}}{\mathtt{G}^T})^2 \cdot \mathtt{N} \cdot \mathtt{P}\big)$ and
$\mathcal{O}\big((\frac{\mathtt{D}}{\mathtt{G}^T})^2\big)$, respectively. Therefore, the time and space complexity of a group TCN layer with $\mathtt{G}^T$ groups is $\mathcal{O}(\frac{\mathtt{D}^2}{\mathtt{G}^T} \cdot \mathtt{N} \cdot \mathtt{P})$ and
$\mathcal{O}(\frac{\mathtt{D}^2}{\mathtt{G}^T})$, respectively, which is $\frac{1}{\mathtt{G}^T}$ of the corresponding standard TCN. For example, in Figure~\ref{fig:tcn}(b), $\mathtt{D}$ = 4, input and output feature maps are split into $\mathtt{G}^T$ = 2 groups, and each group consists of $\mathtt{D}/\mathtt{G}^T$ = 2 feature maps. The number of convolution filters is consequently reduced from $\mathtt{D}^2$ = 16 to $\mathtt{D}^2/\mathtt{G}^T$ = 8.

One drawback of the naive grouping strategy is the lack of information exchange among groups. Thus, we propose to use a {\it shuffled grouping} strategy to support communications among feature map groups. 
As depicted in Figure \ref{fig:tcn}(c), shuffling allows group convolutions to obtain input from different groups, with one input feature map contributing to all groups.
In the implementation, we stipulate that the number of feature maps in each input group is divisible by the number of output groups. The shuffling enhances the naive grouping strategy in a simple but effective way to enable inter-group communication without increasing model complexity.

The group number $\mathtt{G}^T$ is a model structure hyperparameter that controls the balance between the lightness of an L-TCN and its capacity to extract temporal information. 
Intuitively, a larger $\mathtt{G}^T$ improves L-TCN's lightness but reduces its capacity.
Therefore, it is important to tune $\mathtt{G}^T$.
First, $\mathtt{G}^T$ belongs to a small set of candidate values because it can only be a factor of the embedding size (e.g., \{2, 4, 8, 16, 32\} for $\mathtt{D}$ = 64).
Therefore, we employ grid search on the small number of candidates to maximize $\mathtt{G}^T$ while maintaining nearly state-of-the-art accuracy.
Alternatively, it is possible to search for an optimal $\mathtt{G}^T$ more efficiently by applying advanced multi-objective hyperparameter optimization approaches~\cite{morales2022survey}, such as multi-objective Bayesian optimization.
The group number tuning discussed here also applies to the grouping techniques we use in other modules.
The effect of varying $\mathtt{G}^T$ is studied empirically in Section~\ref{sssec:impact_group_number}.

Following previous studies ~\cite{wu2019graph,wu2021autocts}, we adopt the gating mechanism to decide the ratio of temporal information extracted by a shuffled group TCN layer to flow through the model.
Thus, an L-TCN layer is given as follows.
\begin{equation*}
\operatorname{L-TCN}(H) = \operatorname{tanh}(\operatorname{SGTCN}_{{o}}(H \mid \mathtt{G}^T)) \odot \sigma(\operatorname{SGTCN}_{{g}}(H \mid \mathtt{G}^T)),
\end{equation*}
where $\operatorname{SGTCN}_{{o}}$ and $\operatorname{SGTCN}_{{g}}$ are two parallel \emph{shuffled group} TCN branches: the former extracts temporal features and the latter controls the ratio at which the features are passed along. The gating ratio is achieved by the sigmoid function $\sigma(\cdot)$ and is applied to every temporal feature element by the element-wise product $\odot$.

To sum up, an L-TCN layer reduces the time and space complexities to $\frac{1}{\mathtt{G}^T}$ of the standard TCN's counterpart, i.e., to $\mathcal{O}(\frac{\mathtt{D}^2}{\mathtt{G}^T} \cdot \mathtt{N} \cdot\mathtt{P})$ and $\mathcal{O}(\frac{\mathtt{D}^2}{\mathtt{G}^T})$, respectively.
So far, the output feature map is of size $\mathtt{N} \times \mathtt{P} \times \mathtt{D}$. We proceed to present a last-shot compression scheme to reduce the input size for subsequent S-operators.

\subsection{Last-shot Compression}
\label{ssec:last_shot}

\begin{figure}
	\begin{center}
		\includegraphics[width=1\linewidth]{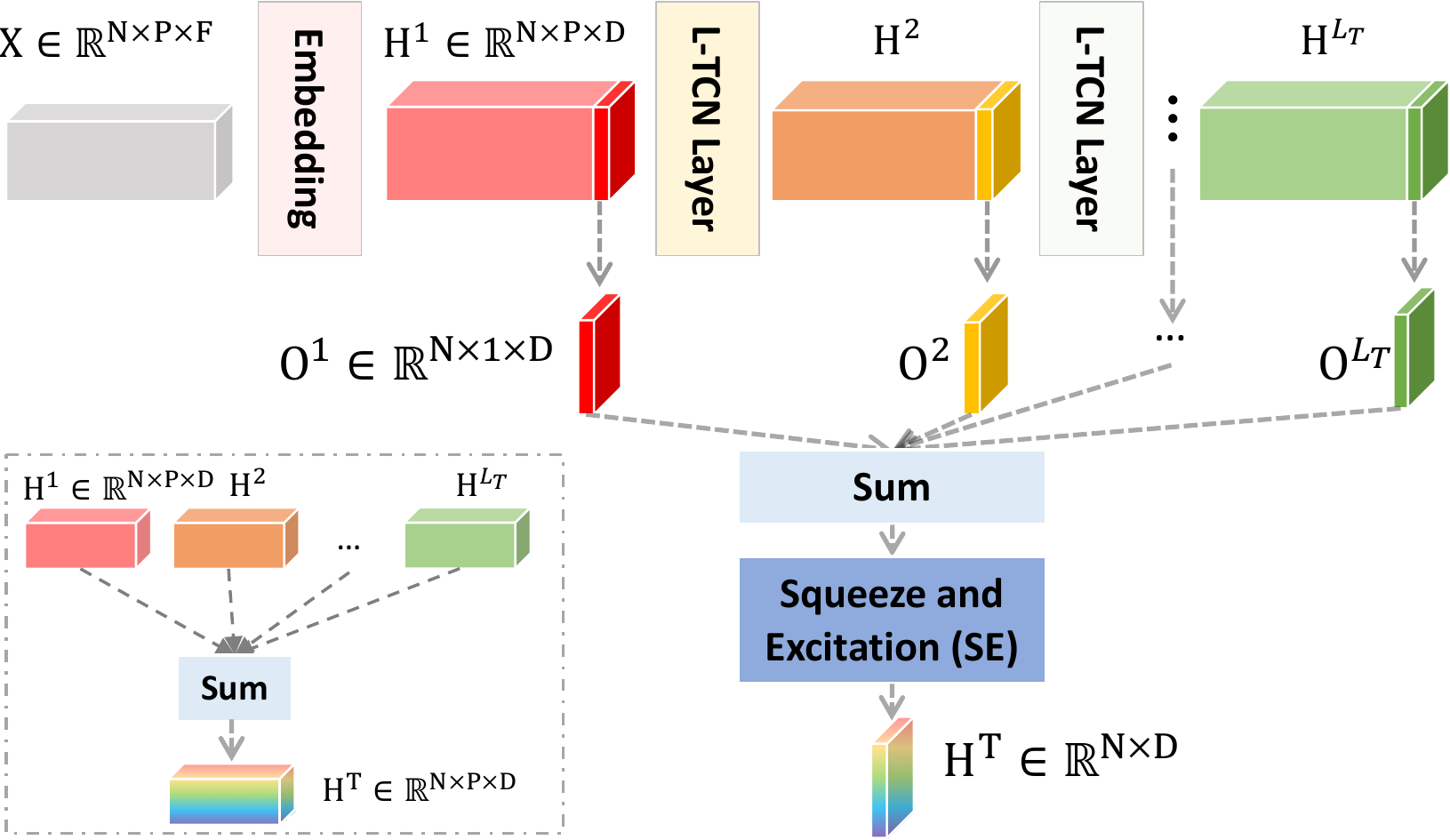}
	\end{center}
	\caption{Feature aggregation after last-shot compression vs the classical feature aggregation~\cite{wu2021autocts} (bottom left).}
	\label{fig:lastshot}
 \vspace{-0.5cm}
\end{figure}

Inspired by the success of residual connections \cite{resnet}, popular CTS models aggregate the features from every ST-block to get the output.
The bottom left corner of Figure \ref{fig:lastshot} illustrates how such a classical aggregation scheme is applied to the L-TCN.
In particular, the features extracted by each L-TCN layer are summed to obtain an aggregated feature tensor of shape $\mathtt{N} \times \mathtt{P} \times \mathtt{D}$, with the temporal dimension $\mathtt{P}$ preserved for subsequent spatial feature extraction.
If we are able to compress the temporal features extracted by each L-TCN layer, the computational overheads of the following S-operators and the aggregation and output module will be reduced.
Note that these two components are both costly (see Table~\ref{tab:distributions}).

To achieve this, we propose a simple but effective mechanism, named \emph{last-shot compression}. The idea comes from the intuition that more distant temporal features are less important for the forecast at the current moment~\cite{cao2003support,tay2002modified,de1999sequential}.
To put it simply, last-shot compression retains only the features at the most recent time step as the output of each L-TCN layer.
An illustration is given in Figure~\ref{fig:tcn} (left), where the rightmost snippets of each layer's output (corresponding to the last time step) are extracted and summed as the input to the subsequent components.

One concern may be that such aggressive compression will completely lose the information from the previous ($\mathtt{P}-1$) time steps of the input feature map.
This is not so because the stacking of dilated convolutions by L-TCNs ensures that the last time step feature of each layer preserves information from several most recent input time steps at different ranges.
For example, in Figure \ref{fig:tcn} (left), the last time step feature from Layer 1 captures the 7th and 8th input time steps, while that from Layer 2 captures the 5th to the 8th input steps, and that from Layer 3 captures all eight input time steps.
Hence, if we aggregate only the last time step feature of each L-TCN layer, we still perceive the full input, but focus more on recent input time steps while performing higher compression on more distant input time steps.
The ablation study in Section~\ref{ssec:ablation} shows that this last-shot compression achieves impressive cost reductions while maintaining model accuracy.

The last-shot compression and the following feature aggregation are illustrated in Figure~\ref{fig:lastshot}.
Let $H^b \in \mathbb{R}^{\mathtt{N} \times \mathtt{P} \times \mathtt{D}}$ be the output feature of the $b$-th ($1 \leq b \leq \mathtt{L}_T$) L-TCN layer, and let $O^b = H^b[:,\mathtt{P}-1,:]$ $ \in \mathbb{R}^{\mathtt{N} \times 1\times \mathtt{D}}$ be the last-step feature of $H^b$.
The aggregation sums the last-step features of all layers, i.e.,
$H = \sum\nolimits_{b=1}^{\mathtt{L}_T} O^b$.
The aggregated feature $H$ is then sent to a squeeze and excitation (SE) module~\cite{hu2018squeeze} for attentive feature representation:
\begin{equation}\label{equation:temporal_feature}
H^{T} = H \cdot \sigma(\text{W}_{s2} \cdot \operatorname{ReLU}(\text{W}_{s1} \cdot H^{\circ})),
\end{equation}
where $H^{\circ}=\operatorname{GlobalAvgPool}(H)\in\mathbb{R}^\mathtt{D}$ is achieved through the global average pooling~\cite{hu2018squeeze}, and $H^T \in \mathbb{R}^{\mathtt{N}\times \mathtt{D}}$ is the final temporal feature. 
Given the reduction ratio $r$ in the SE module, $\text{W}_{s1} \in \mathbb{R}^{\frac{\mathtt{D}}{r} \times \mathtt{D}}$ and $\text{W}_{s2} \in \mathbb{R}^{\mathtt{D} \times \frac{\mathtt{D}}{r}}$ are weight matrices to squeeze the representation that in turn is rescaled back to produce attentions with a sigmoid function $\sigma(\cdot)$ over the original, aggregated feature $H$.


Compared to the classical aggregation scheme that leads to the feature size of $\mathtt{N} \times \mathtt{P} \times \mathtt{D}$, the last-shot compression achieves the feature size of $\mathtt{N} \times \mathtt{D}$ with the temporal dimension having been flattened from $\mathtt{P}$ to 1.
Despite this reduction, the gradually enlarged receptive field of L-TCN (see Section~\ref{ssec:bgtcn}) ensures that the last-shot compression upon L-TCN can obtain the temporal features across all time steps. This property enables further extraction of spatial correlations over time.
The space and time complexities of the subsequent spatial feature extraction are thus reduced by a factor of $1 / \mathtt{P}$.

\subsection{GL-Former}
\label{ssec:ltransformer}
\subsubsection{Background and Overall Design}

Transformers occur in many state-of-the-art CTS forecasting models~\cite{xu2020spatial,park2020st,zhou2021informer,wu2021autocts}.
Aiming for light yet effective S-operators, we propose a GlobalLocal TransFormer (GL-Former) module that aims to extract both global-scale and local-scale spatial correlations among different time series. This way, GL-Former eliminates the exhaustive global-scale spatial correlation extraction seen in the standard Transformer~\cite{vaswani2017attention}.
We proceed to give preliminaries of the standard Transformer and then detail the GL-Former design that targets accuracy and lightness.

A standard Transformer consists of an \emph{encoding layer} and $\mathtt{L}_s$ \emph{attention blocks}.
Given the input $H^{T} \in \mathbb{R}^{\mathtt{N} \times \mathtt{D}}$ generated by the last-shot compression scheme (see Equation~\ref{equation:temporal_feature}), a \emph{positional encoding mechanism} (PE)~\cite{vaswani2017attention} is introduced to the encoding layer. The reason is that by default, Transformers' self-attention operations are unable to interpret the permutations of the input nodes.
Specifically, the encoding layer converts $H^{T}$ to a learned positional encoding $H^\text{PE}$ by incorporating the identity information of CTS nodes:
\begin{equation*}
H^\text{PE} = H^{T}+\text{W}^\text{PE},
\end{equation*}
where $\text{W}^\text{PE} \in \mathbb{R}^{\mathtt{N} \times \mathtt{D}}$ is a learnable matrix capturing the identity information.
The resulting encoding $H^\text{PE}$ is fed to sequential attention blocks, each of which consists of a \emph{multi-head attention} (MHA) layer and a \emph{feed-forward network} (FFN) layer.

An \textbf{MHA layer} is a concatenation of $\mathtt{h}$ repeated self-attention modules (heads) in parallel: 

\begin{equation}\label{equation:mha}
\operatorname{MHA}(H^\text{PE})=\operatorname{concat}(\{ \operatorname{head}_i(H^\text{PE})\}_{i = 1}^ {\mathtt{h}})
\end{equation}
\begin{equation}\label{equation:mha_head}
\operatorname{head}_i(H^\text{PE})= \operatorname{softmax}\big( H^\text{I} \big) \cdot {V}_i
\end{equation}
\begin{equation}\label{equation:mha_global}
H^\text{I} = ({{Q}_i \cdot {K}_i^\mathsf{T}}) /{\sqrt{\mathtt{D}/\mathtt{h}}}
\end{equation}
\begin{equation}\label{equation:mha_qkv}
Q_i = H^\text{PE} \cdot \text{W}^{Q}_{i}, K_i = H^\text{PE} \cdot \text{W}^{K}_{i}, V_i = \text{H}^\text{PE} \cdot \text{W}^{V}_{i},
\end{equation}
where all learnable matrices $\text{W}^{*}_{i} \in \mathbb{R}^{\mathtt{D} \times \frac{\mathtt{D}}{\mathtt{h}}}$ are used to convert the embedding size from $\mathtt{D}$ to $\mathtt{D}/\mathtt{h}$.
In Equation~\ref{equation:mha_head}, each head $\operatorname{head}_i(\cdot)$ is a self-attention module that produces attention scores for its input $H^\text{PE}\in\mathbb{R}^{\mathtt{N} \times \mathtt{D}}$ by making the input interact with itself.
MHA provides impressive power to encode multiple relationships.

An \textbf{FFN layer} merges the output $H^\text{MHA} \in \mathbb{R}^{\mathtt{N}\times\mathtt{D}}$ of Equation~\ref{equation:mha} and provides non-linear activations via two fully-connected layers:
\begin{equation}\label{equation:ffn}
\operatorname{FFN}(H^\text{MHA}) = \operatorname{FFN}^{(1)} (H^\text{MHA}) \cdot \text{W}_2+\text{b}_2
\end{equation}
\begin{equation}\label{equation:ffn_1}
\operatorname{FFN}^{(1)}(H^\text{MHA}) = \operatorname{ReLU}(H^\text{MHA} \cdot \text{W}_1+\text{b}_1),
\end{equation}
where $\text{W}_1 \in \mathbb{R}^{\mathtt{D} \times \mathtt{D}'}$, $\text{W}_2 \in \mathbb{R}^{\mathtt{D}' \times \mathtt{D}}$, $\text{b}_1 \in \mathbb{R}^{\mathtt{D}'}$, and $\text{b}_2 \in \mathbb{R}^{\mathtt{D}}$ are learnable matrices and biases.
The two layers first enlarge the embedding size from $\mathtt{D}$ to $\mathtt{D}'$, which is typically a quadruple of $\mathtt{D}$~\cite{vaswani2017attention}, and then scale it back to $\mathtt{D}$.
The feature generated by the ($\mathtt{L}_S$)-th attention block is the final output of the Transformer, i.e., $H^S$.

\begin{figure}[]
	\begin{center}
		\includegraphics[width=1\linewidth]{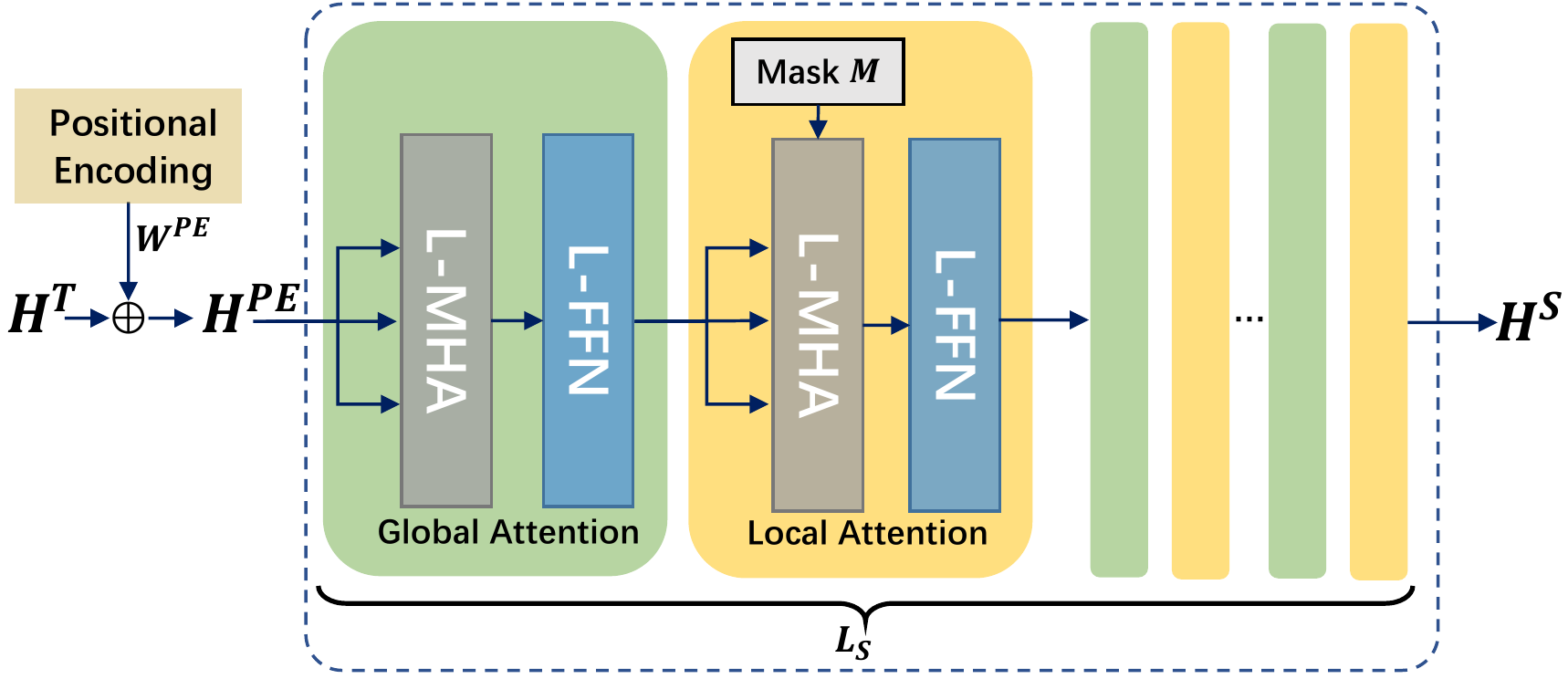}
	\end{center}
	\caption{An example of GL-Former consisting of $\mathtt{L}_S$ alternating global attention blocks and local attention blocks.}
	\label{fig:GL-Former}
 \vspace{-0.5cm}
\end{figure}

A standard attention block introduced above consists of an MHA and an FFN layer and captures global-scale spatial correlations among all CTS nodes.
Although using \emph{global attention blocks} yields powerful feature extraction capabilities, a model that learns such complicated information can be quite hard to train. Injecting prior knowledge into the model is a sensible way to increase model training efficiency, as the model does not need to extrapolate the knowledge from the data itself. Further, the prior knowledge may offer more information beyond the training data, thus helping to regularize the model and prevent overfitting~\cite{vladimirova2019understanding}. Specifically, in our problem setting, using prior knowledge in the data, such as the explicit spatial proximity information of CTS nodes (modeled as an $\mathtt{N}$-by-$\mathtt{N}$ adjacency matrix), we can focus on extracting correlations for only those pairs of nodes that are potentially relevant and can omit computations for other pairs.
This kind of attention block, which we call a \emph{local attention block}, is detailed in Section~\ref{sssec:local_attention}.
As shown in Figure~\ref{fig:GL-Former}, an example GL-Former is a sequence of alternating global and local attention blocks. 
The alternation combats the information loss on local attentions.
Note that the numbers of global and local attention blocks are not necessarily the same. For example, one global attention could be followed by two local attentions.
In addition, we adopt the grouping strategy to ease the computations of the MHA and FFN layers in attention blocks, to be detailed in Section~\ref{sssec：light_mha_ffn}.

\subsubsection{Local Attention Block}\label{sssec:local_attention}
The computation of a local attention follows that of a global attention (Equations~\ref{equation:mha} to~\ref{equation:mha_qkv}), except that a mask function is applied in Equation~\ref{equation:mha_global} to retain pairs of relevant nodes. Specifically, the mask function $\operatorname{mask}(Z, M)$ hides an element $Z[i,j]$ in the feature matrix $Z$ if element $M[i,j]$ of the mask matrix is \texttt{false}; formally,
\begin{equation*}
Z[i,j] = \begin{cases}Z[i,j], & \text{if~}M[i,j] \text{~is~} \texttt{true} \\ -\infty, & \text{otherwise}\end{cases}
\end{equation*}
A hidden value is set to $-\infty$ because the masked feature matrix is sent to a $\operatorname{softmax}(\cdot)$ function (see Equation~\ref{equation:mha_head}).
Besides, a domain may be associated with several adjacency matrices. For example, one adjacency matrix may capture the correlations in terms of the distance between a pair of nodes, while another may capture the correlations in terms of the data dependency between nodes.
In this setting, we obtain mask matrix $M$ by aggregating all relevant adjacency matrices of the domain, i.e., $M = \sum_{i} A_i$, where $A_i$ is one of the adjacency matrices. 
Notably, $A_i$ is a sparse adjacency matrix thresholded by a filtering function~\cite{wu2019graph}, which reduces the impact of noise and makes the model more robust~\cite{chen2022multi}.

With the mask function, the variable $H^\text{I}$ in Equation~\ref{equation:mha_global} is calculated as follows for a local attention block.
\begin{equation}\label{equation:mha_local}
H^\text{I} = \operatorname{mask}\big( ({{Q}_i \cdot {K}_i^\mathsf{T}}) /{\sqrt{\mathtt{D}/\mathtt{h}}}, M \big)
\end{equation}

\subsubsection{L-MHA and L-FNN}\label{sssec：light_mha_ffn}

We propose the light MHA (\textbf{L-MHA}) that adopts a grouping strategy similar to the one proposed for L-TCN in Section~\ref{ssec:ltcn}. 
The input encoding $H^\text{PE}$ of L-MHA is first partitioned into $\mathtt{G}^M$ groups, and then the original MHA (see Equation~\ref{equation:mha}) is applied to each group $H^\text{PE}_j$ ($ 1 \leq j \leq \mathtt{G}^M$). The L-MHA is given as follows.
\begin{equation*}
\begin{aligned}
\operatorname{L-MHA}(H^\text{PE}) & = \operatorname{concat}(\{ \operatorname{MHA}(H^\text{PE}_j)\}_{j = 1}^ {\mathtt{G}^M}), \;\text{where} \\
H^\text{PE}_j & = H^\text{PE}[:, \frac{\mathtt{D} \times (j-1)}{\mathtt{G}^M}:\frac{\mathtt{D} \times j}{\mathtt{G}^M}]
\end{aligned}
\end{equation*}
We apply multi-head attention with the same number $\mathtt{h}$ of heads for each partitioned group. 
The output of $\operatorname{MHA}(H^\text{PE}_j)$ is of size $\mathtt{N} \times \frac{\mathtt{D}}{\mathtt{G}^M}$, and the final output of L-MHA is of size $\mathtt{N} \times \mathtt{D}$, the same as the standard MHA in Equation~\ref{equation:mha}.
Still, the time and space complexities of L-MHA are a fraction ${1}/{\mathtt{G}^M}$ of those of the standard MHA.

Likewise, we implement a light FFN (\textbf{L-FFN}) that partitions the input features into $\mathtt{G}^F$ groups.
We only apply the grouping strategy to the second layer of an original FFN (see Equation~\ref{equation:ffn}) and the computation in the first layer remains as shown in Equation~\ref{equation:ffn_1}.
The first layer $\operatorname{FFN}^{(1)}$ encapsulates the only non-linear activation in an attention block, and lightening it will reduce accuracy considerably.
As a result, L-FFN is processed as follows.
\begin{equation*}
\begin{aligned}
\operatorname{L-FFN}(H^\text{MHA}) & = \operatorname{concat}(\{ \operatorname{FFN}(H^\text{MHA}_j)\})_{j = 1}^ {\mathtt{G}^F}, \;\text{where} \\
H^\text{MHA}_j & = H^\text{MHA}[:,\frac{\mathtt{D} \times (j-1)}{\mathtt{G}^F}:\frac{\mathtt{D} \times j}{\mathtt{G}^F}]
\end{aligned}
\end{equation*}

As only the second FFN layer is lightened, the complexity of L-FFN is ${(1+1/\mathtt{G}^F)}/{(1+1)}$
as the case for the standard FFN counterpart.

\subsection{Assembling \LightCTS{}}
\label{ssec:assemble}
We compose \LightCTS{} using the plain stacking architecture from Figure~\ref{fig:ctsoverview}(b) and using the proposed light T- and S-operator modules (i.e., L-TCN and GL-Former).
In particular, we use one CNN layer to implement the embedding module and configure the aggregation and output module with two fully-connected layers:
\begin{equation*}
\hat{Y} = \operatorname{ReLU}((H^{S}+H^{T}) \cdot \text{W}^o_1+\text{b}^o_1) \cdot\text{W}^o_2+\text{b}^o_2,
\end{equation*}
where $\hat{Y} \in \mathbb{R}^{\mathtt{N} \times \mathtt{L}}$ is the forecast result, $\mathtt{L}=1$ or $\mathtt{Q}$ denotes the forecast dimension depending on whether single-step or multi-step forecasting is performed; $H^{T}$ and $H^{S}$ are the output features of the last-shot compression and GL-Former, respectively; and $\text{W}^o_1$, $\text{W}^o_2$, $\text{b}^o_1$, and $\text{b}^o_2$ are learnable matrices and biases.

Finally, we follow previous work~\cite{rao2022fogs,wu2021autocts,wu2019graph} and employ mean absolute error (MAE) as the loss function.

It should be noted that this simple plain stacking architecture does not lower the effectiveness of feature extraction. The key reason is twofold: 1) the last-shot compression (detailed in Section~\ref{ssec:last_shot}) with the proposed T-operator module L-TCN (detailed in Section~\ref{ssec:ltcn}) are able to emphasize the most recent timestep’s features and to reduce the noise in feature maps, thus improving the subsequent spatial feature extraction; and 2) the proposed S-operator module GL-Former (detailed in Section~\ref{ssec:ltransformer}) addresses both local and global-scale spatial correlations. 
By considering adjacency matrices in the attention block through a mask, GL-Former can capture the prior knowledge of spatial correlations, which the standard Transformer is unable to do, thus enhancing the capability of spatial feature extraction. We report on empirical studies in Section~\ref{ssec:ablation} that offer evidence of the effectiveness of the proposed last-shot compression, GL-Former, and plain stacking.

%% file: content/5.Experiments.tex
\section{Experiments}
\label{sec:experiments}

We evaluate \LightCTS{} on both multi-step and single-step CTS forecasting tasks.
We include many commonly-used benchmark datasets, four for multi-step forecasting (two on traffic flow forecasting and two on traffic speed forecasting) and two for single-step forecasting.
These benchmarks are associated with different accuracy metrics.
To enable direct and fair comparisons, we use the metrics employed in the original papers. 
The code and datasets are made available~\cite{lightcts}.

\subsection{Multi-Step Forecasting}

\subsubsection{Datasets}
\label{sssec:multi_step_datasets}

\begin{itemize}[leftmargin=*]
\item \textbf{PEMS04} \cite{song2020spatial} is a traffic flow dataset collected from 307 sensors on 29 roads in San Francisco Bay Area during January -- March 2018. The traffic flow ranges from 0 to 919 with a mean of 91.74. 

\item \textbf{PEMS08} \cite{song2020spatial} is a traffic flow dataset collected from 170 sensors on 8 roads in San Bernardino during July -- September 2016. The traffic flow ranges from 0 to 1,147 with a mean of 98.17. 

\item \textbf{METR-LA} \cite{li2018diffusion} contains traffic speed data in mph gathered during March -- June 2012 from 207 loop detectors, from the road network of Los Angeles County. The speed ranges from 0 to 70 mph with a mean of 53.98 mph. 

\item \textbf{PEMS-BAY} \cite{li2018diffusion} is a traffic speed dataset gathered from 325 loop detectors in the road network in the San Francisco Bay Area. The speed ranges from 0 to 85.1 mph with a mean of 62.62 mph. 

\end{itemize}
The sampling interval of each dataset is 5 minutes.
The datasets are organized and split (i.e., in train:validation:test) as in previous studies~\cite{wu2019graph,wu2021autocts}. The statistics are summarized in Table ~\ref{tab:datasetmulti}.

\begin{table}[!htbp]
\caption{Dataset statistics for multi-step forecasting.}
\centering
\footnotesize
\resizebox{\linewidth}{!}{
\begin{tabular}{|c|ccccc|c|}
\hline 
Dataset        &Data type        & $\mathtt{N}$  & $\mathtt{T}$  & $\mathtt{P}$ & $\mathtt{Q}$ & Split Ratio \\ \hline\hline
PEMS04         &Traffic flow& 307 & 16,992      & 12    & 12 & 6:2:2           \\
PEMS08         &Traffic flow& 170 & 17,856      & 12    & 12 & 6:2:2           \\ \hline
METR-LA        &Traffic speed& 207 & 34,272     & 12    & 12 & 7:1:2           \\
PEMS-BAY       &Traffic speed& 325 & 52,116     & 12    & 12 & 7:1:2           \\\hline
\end{tabular}
}
\label{tab:datasetmulti}
\end{table}
\subsubsection{Metrics}

We consider accuracy and lightness as follows. 
\begin{itemize}[leftmargin=*]
\item \textbf{Accuracy Metrics}. Following existing multi-step forecasting studies~\cite{li2018diffusion,yu2018spatio,wu2019graph,wu2021autocts}, we use mean absolute error (MAE), root mean squared error (RMSE), and mean absolute percentage error (MAPE) to measure accuracy comprehensively.
The three metrics capture the forecasting accuracy from different perspectives: MAE gives equal weights to all errors, RMSE focuses on the most severe errors, and MAPE highlights the errors when ground truth values are small. Lower MAE, RMSE, and MAPE indicate higher forecasting accuracy.

\item \textbf{Lightness Metrics.} Consistent with the existing conventions \cite{howard2017mobilenets,zhang2018shufflenet} and to eliminate the influence of different DL platforms and operating system conditions (e.g., multiple concurrent running programs), we evaluate the lightness of CTS forecasting models using FLOPs and the number of parameters\footnote{FLOPs and parameter counts are captured by \emph{Fvcore}~\cite{fvcore} from Facebook Research.} considered during \emph{inferencing}. This accords with existing studies~\cite{zhang2018shufflenet,li2021dynamic,tan2019efficientnet}. 
In addition, we report the latency and peak memory use (abbreviated as Peak Mem) of models during inferencing on a low-computational-resource device (see Section~\ref{sssec:implementations}). The results are for practical reference and will vary depending on hardware, software, implementation, and other factors.

\end{itemize}

\subsubsection{CTS Forecasting Models for Comparisons}
All models are implemented using their original code; and if using the same dataset, we report the original results.
\label{sssec:cts_model_multi_step}
\begin{itemize}[leftmargin=*]

    \item \DCRNN{}~\cite{li2018diffusion}. A relatively early DL-based model that adopts diffusion GCNs and GRUs for S- and T-operators, respectively.

    \item \GWNet{}~\cite{wu2019graph}. A widely used benchmark model that integrates adaptive GCNs and TCNs for S- and T-operators, respectively.

    \item \AGCRN{}~\cite{bai2020adaptive}. A comprehensive but costly model that considers dynamic spatial correlations through different time steps.
    
    \item \MTGNN{}~\cite{wu2020connecting}. A successor of \GWNet{} with new graph learning layers and an optimized overall structure.

    \item \AutoCTS{}~\cite{wu2021autocts}. An automated framework that allows heterogeneous ST-blocks with different S/T-operators and their connections through automatic search.

    \item \Enhancenet{}~\cite{cirstea2021enhancenet}. A framework that uses distinct filters for each time series and dynamic adjacency matrices to capture spatial correlations over time. TCNs are used to implement T-operators.

    \item \FOGS{}~\cite{rao2022fogs}. A recent model uses first-order gradients to avoid fitting irregularly-shaped distributions.
    
    \item \AutoCTSKDF{}/\AutoCTSKDP{}. We construct two compressed variants of \AutoCTS{}\footnote{\AutoCTS{} is selected as the teacher model for compression as it generally achieves the highest effectiveness among all competitor methods in our study.} using knowledge distillation (KD) for regression tasks~\cite{takamoto2020efficient}. Specifically, \AutoCTSKDF{} and \AutoCTSKDP{} are compressed to have nearly the same numbers of FLOPs and parameters as \LightCTS{}.

\end{itemize}

\subsubsection{Implementation Details}
\label{sssec:implementations}
All models are trained on a server with an NVIDIA Tesla P100 GPU. To investigate models' inferencing performance in constrained computing environments, we employ an X86 device with a 380 MHz CPU.

The L-TCN has ($\mathtt{L}_T$ = 4) layers with the dilation rate $\delta$ of each layer set to [1, 2, 4, 8] to ensure that the receptive field of the last layer can cover the entire time series length of the input CTS.
The GL-Former has ($\mathtt{L}_S$ = 6) attention blocks for METR-LA and PEMS-BAY, and $\mathtt{L}_S$ = 4 for PEMS04 and PEMS08. The stacking pattern is one global block followed by one local block, as shown in Figure~\ref{fig:GL-Former}. Following parameter tuning, we set the embedding size $\mathtt{D}$ = 48 for METR-LA, $\mathtt{D}$ = 64 for the other three datasets. For all datasets, the group numbers $\mathtt{G}^T$ = 4, $\mathtt{G}^M$ = $\mathtt{G}^F$ = 2, and the reduction ratio $r$ of the SE module is set to 8. 
We adopt the Adam optimizer with a learning rate of 0.002 to train models for 250 epochs.

\subsubsection{Overall Comparisons}
Following existing conventions for direct and fair comparison~\cite{wu2021autocts,rao2022fogs,wu2020connecting}, we report the average accuracy over all 12 future time steps for the PEMS04 and PEMS08 datasets and report the accuracy at the 3rd, 6th, and 12th time steps for the METR-LA and PEMS-BAY datasets.
Tables~\ref{tab:multiresult2} and ~\ref{tab:multiresult} show the results for both the accuracy and lightness measures. In this section, the best results are \textbf{in bold}, and the second-best results are \underline{underlined}. 

Considering {\bf accuracy} on PEMS04 and PEMS08 datasets, Table \ref{tab:multiresult2} shows that \LightCTS{} achieves the best MAE and RMSE results and ranks second regarding MAPE.
Although \FOGS{} achieves the best MAPE on both datasets, it is only marginally better than \LightCTS{} (less than 0.1\%), and its MAE and RMSE rank only around 4th among all models.
The accuracy results on METR-LA and PEMS-BAY datasets in Table \ref{tab:multiresult} show that \LightCTS{} achieves the best MAE, at least the second-best MAPE, and a competitive RMSE (top-3 in most cases). While \LightCTS{} does not always rank in the top-3 in terms of RMSE, it is only negligibly below. For example, for the 15th-minute forecast on METR-LA dataset, \LightCTS{} achieves an RMSE of 5.16 to rank 4th, while the top-3 models achieve 5.11, 5.14, and 5.15; the maximum margin is only 0.05, corresponding to a speed of 0.05 mph traffic speed. Given that there are always fluctuations across real-world datasets, such a small performance difference is insignificant.

%

Considering {\bf lightness}, 
Tables~\ref{tab:multiresult2} and~\ref{tab:multiresult} show that \LightCTS{} has significantly fewer FLOPs and model parameters and achieves lower latency and peak memory use than all other models, with the exception that \AGCRN{} achieves slightly fewer parameters and lower peak memory use on PEMS08 dataset. However, \AGCRN{} is much less accurate. \LightCTS{} clearly uses fewer resources than the two most accurate competing models, \Enhancenet{} and \AutoCTS{} (e.g., less than 1/6 and 1/10 FLOPs), while maintaining comparable accuracy.

Besides, although KD~\cite{takamoto2020efficient} enables creating the \AutoCTSKDF{} and \AutoCTSKDP{} models that have FLOPs and parameter counts comparable to those of \LightCTS{}, their effectiveness metrics are significantly lower, as seen in Tables~\ref{tab:multiresult2} and~\ref{tab:multiresult}.

In summary, \LightCTS{} offers substantially reduced computational and storage overheads while providing top-tier accuracy at multi-step CTS forecasting. \LightCTS{} thus offers unique value because it enables CTS forecasting with no accuracy penalty using limited resources, which are often found in real-world applications. 
In addition, it also lowers costs when deployed in non-constrained settings, such as servers.

\begin{table}[ht]

\caption{Accuracy and lightness comparison for multi-step traffic flow forecasting.}
\resizebox{\linewidth}{!}{

\begin{tabular}{c|c||cccc||ccc}
\toprule
Data & Models & \begin{tabular}[c]{@{}c@{}}FLOPs\\ (unit: M)\end{tabular} & \begin{tabular}[c]{@{}c@{}}Params\\ (unit: K)\end{tabular}& \begin{tabular}[c]{@{}c@{}}Latency\\ (unit: s)\end{tabular}& \begin{tabular}[c]{@{}c@{}}Peak Mem\\ (unit: Mb)\end{tabular}& MAE & RMSE & MAPE  \\ \midrule
\multirow{11}{*}{\rotatebox{90}{PEMS04}} &  \DCRNN{} & 3739&371 & 22.9 & 8.1 & 24.70 & 38.12 & 17.12\%  \\
                        & \GWNet{}    & 1277 & 311  & 4.8 & 6.8& 19.16 & 30.46 & 13.26\%  \\
                        & \AGCRN{}    & 3936 & 749  & 9.5 & 19.2 & 19.83 & 32.26 & 12.97\%  \\
                        & \MTGNN{}    & \underline{393} & 547 & 5.3 & 12.1 & 19.32 & 31.57 & 13.52\% \\
                        & \AutoCTS{}  & 2043 & 368 & 5.2 & 7.8 & 19.13 & {30.44} & {12.89\%}  \\
                        & \Enhancenet{}  & 969 & \underline{283} & \underline{4.2} & \underline{6.0} & \underline{19.11} & \underline{30.34} & 14.33\%  \\
                        & \FOGS{}       & 5936 & 2366  & 14.6 & 42.9& 19.34 & {31.20} & \textbf{12.71\%}  \\ \cmidrule{2-9}
                        & \AutoCTSKDF{}  & 196 & 17  &  3.0 & 5.3 & 25.17 &  38.85& 17.06\%    \\ 
                        & \AutoCTSKDP{}  & 1278 & 186 &  4.4 & 6.8 & 22.83 &  35.21& 15.55\%    \\ \cmidrule{2-9}
                        & \textbf{\LightCTS{}} & \textbf{147}& \textbf{185} & \textbf{1.1} & \textbf{4.7} &  \textbf{18.79} & \textbf{30.14} & \underline{12.80\%}  \\ \midrule
\multirow{11}{*}{\rotatebox{90}{PEMS08}} & \DCRNN{}   & 2070 & 371 & 14.9 &  4.8 & 17.86 & 27.83 & 11.45\%  \\
                        & \GWNet{}    & 479 & 309& \underline{1.0}& 4.0   & 15.13 & 24.07 & 10.10\%  \\
                        & \AGCRN{}    & 726 & \textbf{150}  & 3.2& \textbf{2.6} & 15.95 & 25.22 & 10.09\%  \\
                        & \MTGNN{}    & \underline{153} & 352 & 2.7  & 4.6  & 15.71 & 24.62 & 10.03\% \\
                        & \AutoCTS{}  & 808 & 366   & 3.7 & 4.7& \underline{14.82} & 23.64 & {9.51\%}  \\
                        & \Enhancenet{} & 365 & 275 & 1.7 & 3.8  & \underline{14.82} & \underline{23.60} & 9.58\%   \\
                        & \FOGS{}       & 1949 & 1294 & 6.6& 14.2  & 14.92 & 24.09 & \textbf{9.42\%}   \\ \cmidrule{2-9}
                        & \AutoCTSKDF{}  & 100 & 22 & 1.3  &  3.3& 19.47 & 30.26 & 13.77\%    \\ 
                        & \AutoCTSKDP{}  & 485 &  181  & 1.9 &  4.0& 16.89 & 26.40 & 11.08\%    \\ \cmidrule{2-9} 

                        & \textbf{\LightCTS{}} & \textbf{70} & \underline{177}& \textbf{0.4}  & \underline{2.8}  & \textbf{14.63} & \textbf{23.49} & \underline{9.43\%}   \\ \bottomrule
\end{tabular}
}
\label{tab:multiresult2}
\end{table}

\begin{table*}[ht]
\footnotesize
\centering
\caption{Accuracy and lightness comparison for multi-step traffic speed forecasting.}
\begin{tabular}{c|l||cccc||ccc|ccc|ccc}
\toprule
Data &
  \multicolumn{1}{c||}{Model} &
    FLOPs &
    Params &
    Latency &
    Peak Mem &
  \multicolumn{3}{c|}{15 mins} &
  \multicolumn{3}{c|}{30 mins} &
  \multicolumn{3}{c}{60 mins} \\ 
\multirow{11}{*}{\rotatebox{90}{METR-LA \;\;\;\;\;\;\;\;\;}} & 
    &
    (unit: M)  & (unit: K)& (unit: s)  & (unit: Mb)&
  MAE &
  RMSE &
  MAPE&
   MAE &
  RMSE &
  MAPE&
   MAE &
  RMSE &
  MAPE
  \\ \midrule
 &
 \DCRNN{} &
  2521&
  436&
  16.4&
  13.6&
  2.77 &
  5.38 &
  7.30\% &
  3.15 &
  6.45 &
  8.80\% &
  3.60 &
  7.60 &
  10.50\%\\
 &
  \GWNet{} &
  658&
  \underline{309}&
    1.7&
  8.9&
  \underline{2.69} &
  {5.15} &
  {6.90\%} &
  3.07 &
  6.22 &
  8.37\% &
  3.53 &
  7.37 &
  10.01\% \\
 &
  
  \AGCRN{} &
  2453&
  748&
  7.5&
  22.6&
  2.83 &
  5.45 &
  7.56\% &
  3.20 &
  6.55 &
  8.79\% &
  3.58 &
  7.41 &
  10.13\% \\
 &
  \MTGNN{} &
  \underline{208}&
  405&
  3.9&
  11.9&
  \underline{2.69} &
  5.18 &
  {6.86}\% &
  {3.05} &
  {6.17} &
  {8.19\%} &
  {3.49} &
  {7.23} &
  {9.87\%} \\
 &
  \AutoCTS{} &
  1090&
  366&
  2.8&
  11.4&
  \textbf{2.67} &
  \textbf{5.11} &
  \textbf{6.80\%} &
  \underline{3.05} &
  \underline{6.11} &
  \underline{8.15\%} &
  \underline{3.47} &
  \textbf{7.14} &
  \underline{9.81\%}\\
 &
    \Enhancenet{} &
  648&
  {453}&
  2.6&
  13.4&
  \underline{2.69} &
  \underline{5.14} &
  {6.93\%} &
  {3.06} &
  \textbf{6.10} &
  {8.29\%} &
  {3.49} &
  {7.23} &
  {9.96\%} \\
 &
     \FOGS{} &
   2858&
  1524&
  7.4&
  45.9&
  2.72 &
  5.20 &
  7.05\% &
  3.12 &
  6.30 &
  8.60\%&
  3.64 &
  7.61 &
  10.62\% \\ \cmidrule{2-15}
&
\AutoCTSKDF{} &
  95
  &
15  &
  {1.6}&
  {5.9}&
  3.04 &
  5.80&
  8.49\%&
  3.57&
   7.03&
  10.49\%&
  4.19&
   8.34&
   12.73\%
  \\
   &
\AutoCTSKDP{} & 
  595
  &
 155
  &
  2.2&
  7.1&  
2.78 &
  5.21&
  7.33\%&
  3.18&
   6.23&
  9.00\%&
  3.64&
   7.28&
   10.97\%
  \\ \cmidrule{2-15}
   &
  \textbf{\LightCTS{}} &
  \textbf{71}&
  \textbf{133}&
  \textbf{0.3}&
  \textbf{5.6}&
  \textbf{2.67} &
  {5.16} &
  \underline{6.82\%} &
  \textbf{3.03} &
  {6.16} &
  \textbf{8.11\%} &
  \textbf{3.42} &
  \underline{7.21} &
  \textbf{9.46\%} \\ \midrule
\multirow{11}{*}{  \rotatebox{90}{PEMS-BAY}} &

\DCRNN{} &
  5386&
  436&
  22.6&
  13.8&
  1.38 &
  2.95 &
  2.90\% &
  1.74&
  3.97 &
  3.90\% &
  2.07 &
  4.74 &
  4.90\%\\
 &
  \GWNet{} &
  1408&
  \underline{312}&
  \underline{3.7}&
  \underline{10.3}&
  \textbf{1.30} &
  \underline{2.74} &
  2.73\% &
  \underline{1.63} &
  3.70 &
  3.67\% &
  1.95 &
  4.52 &
  4.63\% \\
 &
  \AGCRN{} &
  4224&
  749&
  10.1&
  22.7&
  1.35 &
  2.83 &
  2.87\% &
  1.69 &
  3.81 &
  3.84\% &
  1.96 &
  4.52 &
  4.67\% \\&
  \MTGNN{} &
  \underline{432}&
  573&
  7.6&
  19.2&
  \underline{1.32} &
  2.79 &
  2.77\% &
  1.65 &
  3.74 &
  3.69\% &
  1.94 &
  4.49 &
  4.53\% \\
 &
  \AutoCTS{} &
  2295&
  369&
  5.9&
  11.9&
  \textbf{1.30} &
  \textbf{2.71} &
  \textbf{2.69\%} &
  \textbf{1.61} &
  \textbf{3.62} &
  \textbf{3.55\%} &
  \textbf{1.89} &
  \textbf{4.32} &
  \textbf{4.36\%} \\
 &
     \Enhancenet{} &
  1442&
  474&
  5.4&
  14.2&
  1.33 &
  2.81 &
  2.80\% &
  1.64 &
  3.72 &
  3.65\% &
  \underline{1.93} &
  \underline{4.47} &
  4.51\% \\
 &
      \FOGS{} &
   6608&
  {2551}&
  16.3&
  76.5&
  {1.38} &
  {2.91} &
  {2.94\%} &
  {1.73} &
  3.93 &
  3.97\%&
  2.09 &
  4.71 &
  4.96\%\\ \cmidrule{2-15}
 &
\AutoCTSKDF{} &
  218&
  18&
   {3.4}&
  {9.5}&
  1.42 &
   2.92&
   2.95\%&
   1.78&
      4.02&
   3.99\%&
   2.11&
   4.82&
   5.06\%
   \\
 &
  \AutoCTSKDP{} &
  1431&
  248&
   4.4&
  10.4&
  1.38 &
   2.85&
   2.87\%&
   1.72&
      3.82&
   3.86\%&
   2.06&
   4.62&
   4.73\%
   \\  \cmidrule{2-15}
 &
  \textbf{\LightCTS{}} &
  \textbf{208}&
  \textbf{236}&
  \textbf{1.2}&
  \textbf{9.2}
  &
  \textbf{1.30} &
  {2.75} &
  \underline{2.71\%} &
  \textbf{1.61} &
  \underline{3.65} &
  \underline{3.59\%} &
  \textbf{1.89} &
  \textbf{4.32} &
  \underline{4.39\%}\\ \bottomrule
\end{tabular}

\label{tab:multiresult}
\end{table*}

\begin{table*}[ht]
\caption{{Accuracy and lightness comparison for single-step CTS forecasting.}}
\centering
\footnotesize
\begin{tabular}{c|l||cccc||cc|cc|cc|cc}
\toprule
Data &
  \multicolumn{1}{c||}{Model} &
  FLOPs &
  Params&
  Latency&
Peak Mem &

  \multicolumn{2}{c|}{3-th} &
  \multicolumn{2}{c|}{6-th} &
  \multicolumn{2}{c|}{12-th} &
  \multicolumn{2}{c}{24-th}  \\
 &
   &
  (unit: M) &
  (unit: K)&
  (unit: s) &
  (unit: Mb)&
  RRSE &
  CORR &
  RRSE &
  CORR &
  RRSE &
  CORR &
  RRSE &
  CORR  \\ \midrule
\multirow{8}{*}{\rotatebox{90}{Solar-Energy}} &
  \DSANet{} &
  914 &
  6377&
  0.8&
  32.5&
  0.1822 &
  0.9842 &
  0.2450 &
  0.9701 &
  0.3287 &
  0.9444 &
  0.4389&
  0.8943  \\
 &
 \MTGNN{} &
  {1090} &
  348&
  0.5&
  9.9&
  0.1778 &
  {0.9852} &
  {0.2348} &
  {0.9726} &
  0.3109 &
  0.9509 &
  0.4270&
  0.9031  \\
 &
  \MAGNN{} &
  \underline{492} &
  105 &
  0.4&
   9.2&
 {0.1771} &
  \underline{0.9853} &
  0.2361 &
  0.9724 &
  {0.3105} &
  {0.9539} &
  \textbf{0.4108} &
  \textbf{0.9097} \\
 &
  \AutoCTS{} &
  2237 &
  \underline{91}&
  1.1&
  17.6&
  \underline{0.1750} &
  \underline{0.9853} &
  \underline{0.2298} &
  \underline{0.9763} &
  \underline{0.2957} &
  \underline{0.9566} &
  {0.4143} &
  \textbf{0.9097}  \\   \cmidrule{2-14}

&
 \AutoCTSKDF{} &
   418&
  12&
  0.4&
  9.2&
 0.1802  &
 0.9834&
 0.2463&
 0.9696&
 0.3332&
 0.9403&
 0.4277&
 0.9021

  \\
&
  \AutoCTSKDP{} &
   1196&
  41&
  0.7&
   13.4&
0.1785  &
 0.9844&
 0.2371&
 0.9736&
 0.3288&
 0.9435&
 0.4196&
 0.9043

  \\ \cmidrule{2-14}

 &
  \textbf{\LightCTS{}} &
   \textbf{169} &
  \textbf{38} &
  \textbf{0.2}&
  \textbf{8.6}&
  \textbf{0.1714} &
  \textbf{0.9864} &
  \textbf{0.2202} &
  \textbf{0.9765} &
  \textbf{0.2955} &
  \textbf{0.9568} &
  \underline{0.4129} &
  \underline{0.9084} \\ \midrule
\multirow{8}{*}{\rotatebox{90}{Electricity}} &
  
  \DSANet{} &
  2262 &
  6377 &
  1.2&
  53.9&
  0.0855 &
  0.9264 &
  0.0963 &
  0.9040 &
  0.1020 &
  0.8910 &
  0.1044 &
  0.8898  \\
 &
  \MTGNN{} &
  {4800} &
  362&
  1.5&
  21.4&
  {0.0745} &
  0.9474 &
  0.0878 &
  0.9316 &
  {0.0916} &
  \underline{0.9278} &
  {0.0953} &
  \underline{0.9234}  \\
 &
  \MAGNN{} &
  \underline{2215} &
  120&
  0.8&
  20.3&
  {0.0745} &
  \underline{0.9476} &
  {0.0876} &
  \underline{0.9323} &
  \underline{0.0908} &
  \textbf{0.9282} &
  0.0963 &
  0.9217  \\
 &
  \AutoCTS{} &
  8740 &
 \underline{95} &
 3.2&
  21.3&
  \underline{0.0743} &
  \textbf{0.9477} &
  \underline{0.0865} &
  0.9315 &
  0.0932 &
  0.9247 &
  \textbf{0.0947} &
  \textbf{0.9239} \\  \cmidrule{2-14}
  &

\AutoCTSKDF{} & 
   1858&
  16&
  1.8&
  12.4&

0.0818&
0.9292&
0.0949&
0.9148&
0.1003&
0.9007&
0.1018&
0.8935  \\

 &

  \AutoCTSKDP{} & 
   3937&
  33&
  2.3&
  17.3&
0.0764&
0.9442&
0.0899&
0.9275&
0.0934&
0.9188&
0.0983&
0.9071  \\   \cmidrule{2-14}
  
 &

  
  \textbf{\LightCTS{}} &
  \textbf{239} &
  \textbf{27}&
  \textbf{0.4}&
  \textbf{10.0}&
  \textbf{0.0736} &
  0.9445 &
  
  \textbf{0.0831} &
  \textbf{0.9343} &
  
  \textbf{0.0898} &
  0.9261 &
  \underline{0.0952} &
  0.9215  \\ \bottomrule
\end{tabular}
\label{tab:singleresult}

\end{table*}

\subsection{Single-Step Forecasting}

\subsubsection{Datasets}\label{sssec:single_step_datasets}
\begin{itemize}[leftmargin=*]
\item \textbf{Solar-Energy} \cite{lai2018modeling} contains records of solar power production in megawatt-hour (MWh) collected from 137 photovoltaic plants in Alabama during 2006. The records range from 0 to 88.9 MWh with a mean of 6.4 MWh. The sampling interval is 10 minutes.

\item\textbf{Electricity} \cite{lai2018modeling} contains records of electricity consumption in kilowatt-hour (kWh) for 321 clients in Portugal during 2012 -- 2014. The values range from 0 to 764,000 kWh, with an average of 2,514 kWh. The sampling interval is 15 minutes.
\end{itemize}
The data organization and data splitting follow existing work~\cite{wu2020connecting,wu2021autocts}. Dataset statistics are summarized in Table~\ref{tab:datasetsingle}.

\begin{table}[H]
\caption{Dataset statistics for single-step forecasting.}
\centering
\begin{tabular}{|c|cccc|c|}
\hline
Dataset        & $\mathtt{N}$  & $\mathtt{T}$ & $\mathtt{P}$ & $\mathtt{Q}$ & Split Ratio \\ \hline\hline 
Solar-Energy   & 137 & 52,560      & 168   & \{3, 6, 12, 24\}  & 6:2:2          \\
Electricity    & 321 & 26,304     & 168   &\{3, 6, 12, 24\}& 6:2:2           \\ \hline
\end{tabular}
\label{tab:datasetsingle}
\end{table}

\subsubsection{Metrics}
We use the same lightness metrics as the multi-step forecasting task.
We use root relative squared error (RRSE) and correlation coefficient (CORR) to measure forecasting accuracy, which are the conventional metrics used in single-step CTS forecasting~\cite{wu2021autocts,chen2022multi,chang2018memory}. Specifically, RRSE indicates how well a model performs w.r.t. the average of the true values, while CORR measures the strength of the linear correlation between the forecast results and the true values. The more accurate the model, the lower the RRSE and the higher the CORR.

\subsubsection{CTS Models in Comparisons}
We include two CTS models that are specifically designed for single-step forecasting:
\begin{itemize}[leftmargin=*]
    \item \DSANet{}~\cite{huang2019dsanet}. A dual self-attention network for multivariate time series forecasting.

    \item \MAGNN{}~\cite{chen2022multi}. A  multi-branch model that extracts temporal features at different time scales.
\end{itemize}
In addition, we include \MTGNN{} \cite{wu2020connecting} and \AutoCTS{} \cite{wu2021autocts} (see Section~\ref{sssec:cts_model_multi_step}) as they also support single-step forecasting.
We use the settings that achieve the best accuracy for the comparison models; or if using the same dataset, we report their original results.

\subsubsection{Implementation Details}
To build \LightCTS{} for single-step forecasting, we do almost the same as for multi-step forecasting.
The differences are as follows.
According to parameter tuning, we set $\mathtt{D}$ = 32 for the Solar-Energy dataset and $\mathtt{D}$ = 24 for the Electricity dataset, and we set the dilation rates in the ($\mathtt{L}_T$ = 8) L-TCN layers to [1, 2, 4, 8, 16, 32, 48, 64].
We adopt a GL-Former with ($\mathtt{L}_S$ = 2) attention blocks (i.e., one global attention followed by one local attention).
We adopt the Adam optimizer with a learning rate of 0.0005 to train models for 100 epochs.

\subsubsection{Overall Comparisons}
Table~\ref{tab:singleresult} shows the results on Solar-Energy and Electricity datasets. 
In accordance with existing research~\cite{chen2022multi,wu2020connecting}, we report single-step forecasting results for the 3rd, 6th, 12th, and 24th future time steps.


We observe similar trends as for multi-step forecasting. In terms of accuracy, \LightCTS{} achieves the best performance on most of the comparison items, and it is very competitive on the others.
Next, \LightCTS{} is the most lightweight model with far fewer FLOPs and parameters than all competitors. For example, \LightCTS{} uses less than 1/10 (\emph{resp.} 1/3) FLOPs than \AutoCTS{} (\emph{resp.} \MAGNN{}). In addition, \LightCTS{} has the lowest latency and peak memory use among all baselines. 
\LightCTS{}'s low requirement of computing resource means a great potential to be deployed in resource-constrained environments.

\subsection{Parameter Study}
We study systematically the impact of key \LightCTS{} hyperparameters, including the embedding size $\mathtt{D}$, the group number $\mathtt{G}^T$ of L-TCN, and the attention block number $\mathtt{L}_S$ of GL-Former. These hyperparameters are selected as they are adjustable and affect model performance noticeably. We summarize the results in Figures \ref{fig:ms} and \ref{fig:ss} for multi-step forecasting on PEMS08 and single-step forecasting on Solar-Energy, respectively. We report the results on the other datasets in the supplemental material~\cite{lightcts}. 

\subsubsection{Impact of Embedding Size $\mathtt{D}$}

\begin{figure*}[!htbp]
	\begin{center}
		\includegraphics[width=0.85\linewidth]{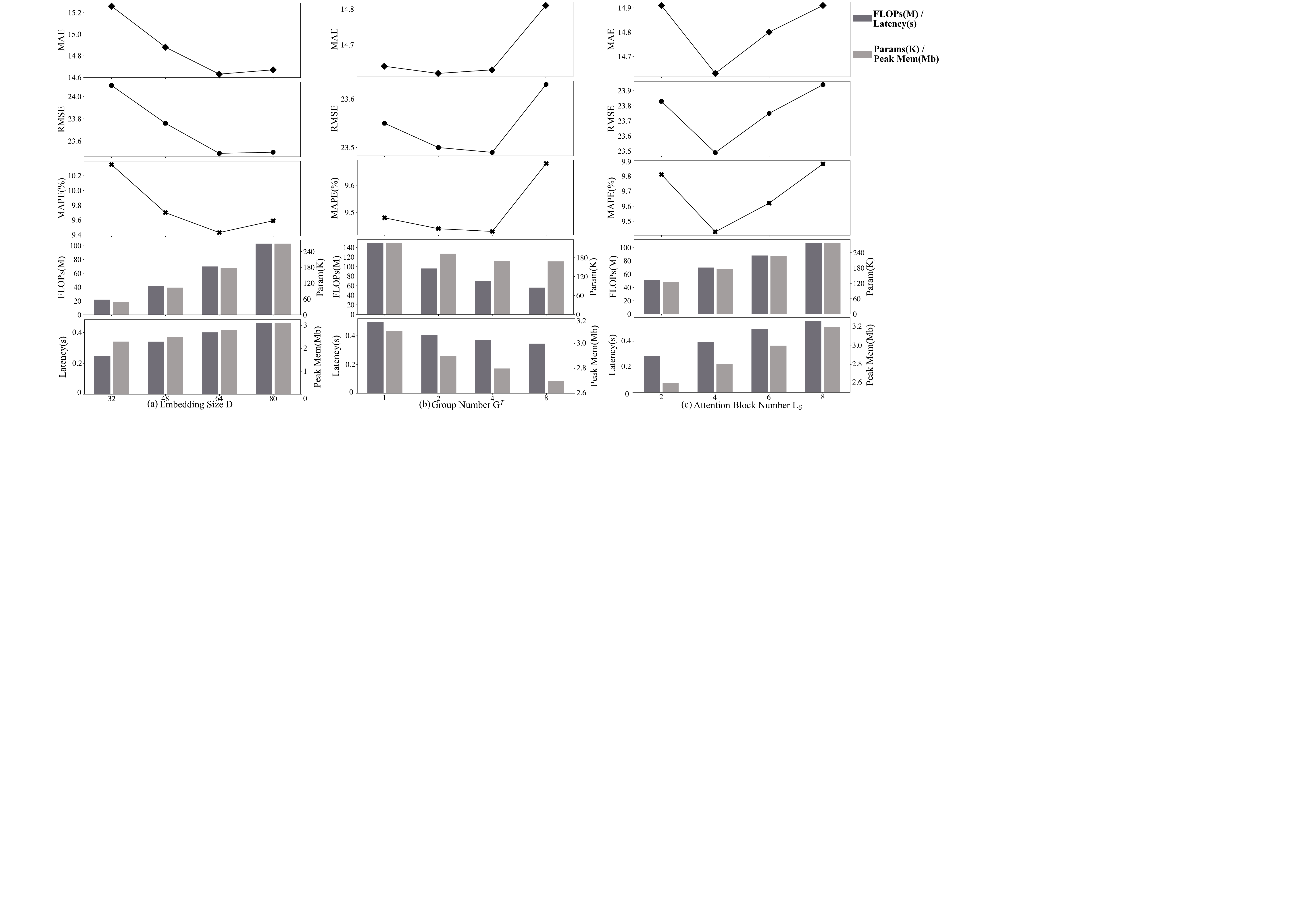}
	\end{center}
	\caption{Impact of (a) embedding size $\mathtt{D}$, (b) group number $\mathtt{G}^T$, and (c) attention block number $\mathtt{L}_S$ in GL-Former for multi-step forecasting on PEMS08 dataset.}
\label{fig:ms}
\end{figure*}

Figure~\ref{fig:ms}(a) shows the impact of $\mathtt{D}$ on model accuracy and lightness of single-step forecasting on the PEMS08 dataset. 
Both the FLOPs and number of parameters increase steadily as $\mathtt{D}$ increases, and so do the latency and peak memory use.

\begin{figure*}[!htbp]
	\begin{center}
		\includegraphics[width=0.85\linewidth]{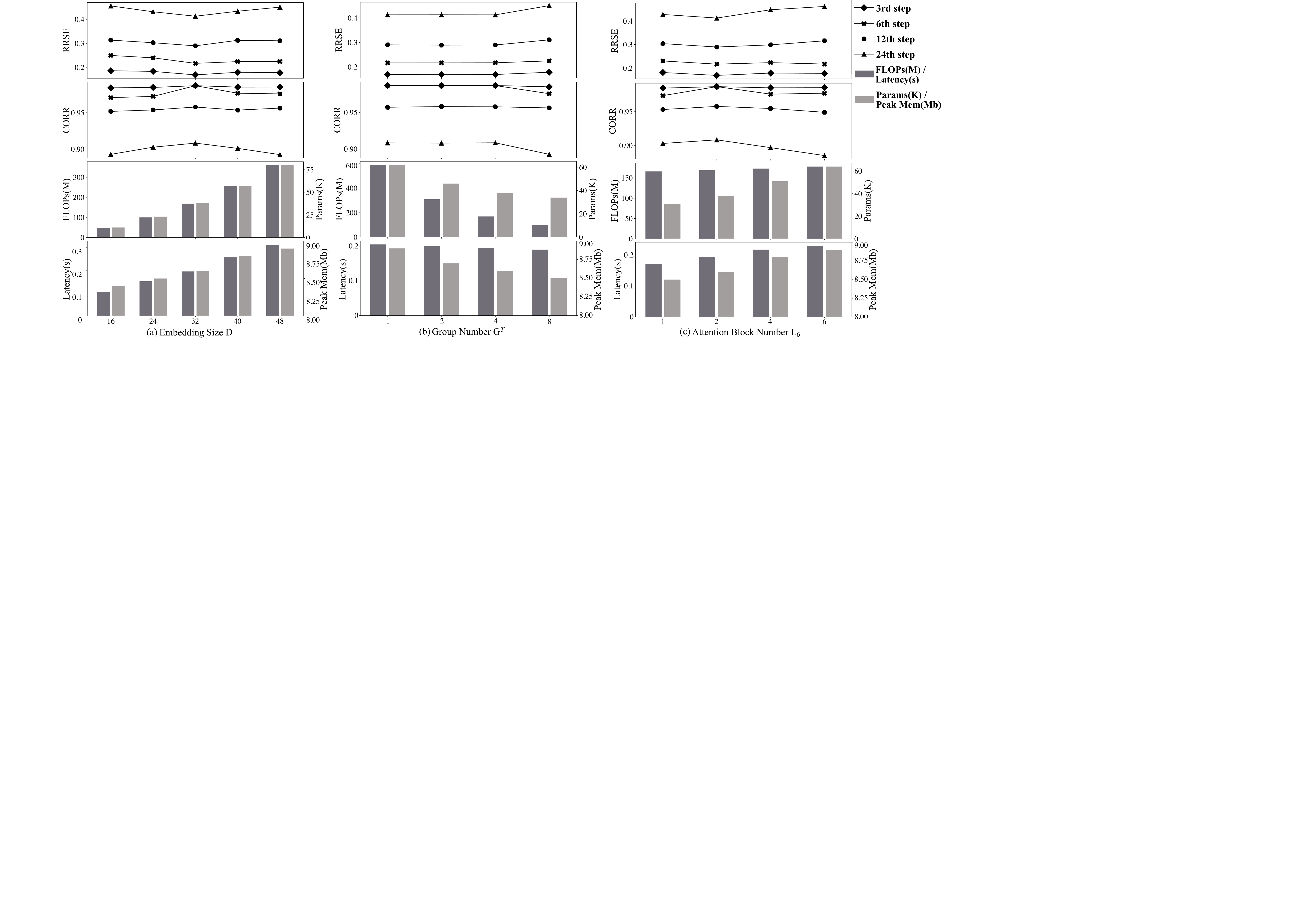}
	\end{center}
	\caption{Impact of (a) embedding size $\mathtt{D}$, (b) group number $\mathtt{G}^T$, and (c) attention block number $\mathtt{L}_S$ in GL-Former for single-step forecasting on Solar-Energy dataset.}
\label{fig:ss}
\end{figure*}

Considering accuracy, as $\mathtt{D}$ grows from 32 to 64, MAE, RMSE, and MAPE decrease moderately. 
However, when $\mathtt{D}$ goes up from 64 to 80, the forecasting errors increase slightly.
The reason may be that a smaller $\mathtt{D}$ restricts the model's ability to extract ST-features, while a larger $\mathtt{D}$ may introduce redundancy into the model and make it difficult to train, and can lead to overfitting.
In Figure~\ref{fig:ss}(a), the single-step forecasting exhibits similar trends as the multi-step counterpart. The FLOPs and number of parameters increase while the forecasting errors drop first and climb up afterwards.

The results are consistent with those of existing studies \cite{wu2020connecting,chen2022multi}---directly cutting the embedding size to a small value inevitably reduces model accuracy. Thus, a new design for manipulating $\mathtt{D}$, such as our L-TCN and GL-Former, is effective at enabling lightweight and accurate CTS forecasting models.

\subsubsection{Impact of Group Number $\mathtt{G}^T$}
\label{sssec:impact_group_number}

We further investigate the impact of the group number $\mathtt{G}^T$ of L-TCN on the performance and lightness of \LightCTS{}. This parameter controls the tradeoff between L-TCN's capacity for temporal information extraction and its lightness.

The multi-step forecasting results in Figure ~\ref{fig:ms}(b) show that when $\mathtt{G}^T$=4, the evaluation errors reach a minimum that is below the value for $\mathtt{G}^T$=1, i.e., the standard TCN without grouping.
This implies that our grouping strategy can remove redundant connections without compromising forecasting accuracy. However, due to information loss, the evaluation errors increase substantially when the group size is increased to 8. 
A similar pattern is seen for single-step forecasting in Figure~\ref{fig:ss}(b), where the best accuracy is also obtained when $\mathtt{G}^T$=4.

Further, additional experiments show that similar patterns appear for the other group number hyperparameters, namely $\mathtt{G}^M$ and $\mathtt{G}^F$. For brevity, we report on these experiments in the supplemental material \cite{lightcts}.

\subsubsection{Impact of Attention Block Number $\mathtt{L}_S$ in GL-Former}

\LightCTS{} supports an elastic number of attention blocks in GL-Former.
Hence, it is of interest to understand how the GL-Former attention block number $\mathtt{L}_S$ affects accuracy and lightness.
The number $\mathtt{L}_T$ of L-TCN layers is also tunable, with its value being relevant to the input time series $\mathtt{P}$ for producing sufficient receptive fields. We thus focus on varying and testing the attention block number $\mathtt{L}_S$.

As shown in Figures~\ref{fig:ms}(c) and ~\ref{fig:ss}(c), for both multi-step and single-step forecasting, as GL-Former goes deeper (i.e., larger $\mathtt{L}_S$), the evaluation errors first decrease and then increase. This happens because although deeper models can theoretically better extract information, they are prone to overfitting or hard to train~\cite{vladimirova2019understanding}. Notably, in Figure~\ref{fig:ss}(c), since the T-operators consume a dominant proportion of FLOPs due to the large input length ($\mathtt{P}$ = 168) of time series, the overall FLOPs and peak memory use of \LightCTS{} increase only slightly when $\mathtt{L}_S$ is increased.

\subsection{Ablation Study}
\label{ssec:ablation}

\begin{table}[]
\centering
\caption{Ablation study on PEMS08. Refer to Section \ref{ssec:ablation} for the details of the models.}
\centering
\resizebox{\linewidth}{!}{
\begin{tabular}{lccccccc}
\toprule
Model &
  \begin{tabular}[c]{@{}c@{}}FLOPs\\ (unit: M)\end{tabular} &
  \begin{tabular}[c]{@{}c@{}}Params\\ (unit: K)\end{tabular} &
  \begin{tabular}[c]{@{}c@{}}Latency\\ (unit: s)\end{tabular} &
  \begin{tabular}[c]{@{}c@{}}Peak Mem\\ (unit: Mb)\end{tabular} &

  MAE &
  RMSE &
  MAPE \\ \midrule
\textbf{\LightCTS{}}  & 70  & 177 & 0.4  & 2.8& 14.63 & 23.49 & 9.43\% \\ \midrule
  \LightCTST{} & 149 & 226& 0.5& 3.1& 14.64 & 23.55 & 9.48\%  \\
  \LightCTSLS{} & 390 & 285 & 0.6& 3.7 & 15.31 &  24.21 & 10.55\%  \\ 
  \LightCTSM{} & 75  & 239& 0.5 &3.0 & 14.57 & 23.48 & 9.48\% \\
  \LightCTSF{} & 75  & 238& 0.5 & 3.0& 14.64 & 23.59 & 9.66\% \\
    \LightCTSL{}  &70&177& 0.4& 2.8&14.70& 23.71& 9.55\% \\   \midrule
  
  \LightCTSA{} & 390 & 285& 0.6& 3.7& 15.64 & 24.52 & 10.96\% \\ 
  \LightCTSCGCN{}  &53&74 & 0.3 &1.7&16.53 & 26.34 & 10.63\% \\ 
  \LightCTSDGCN{}  &95&154 &0.4 &2.6 &16.24& 25.79 & 10.72\% \\ 
  \bottomrule
\end{tabular}}
\label{tab:ablation}
\end{table}

We conduct an ablation study on PEMS08 dataset to understand the contribution of each component in \LightCTS{}.
Specifically, we implement a group of \LightCTS{} variants by removing one of the lightweight components and observe the impact on both accuracy and lightness. 
The variants include: 
\begin{itemize}[leftmargin=*]
\item \LightCTST{} that substitutes the L-TCN with the standard TCN as the T-operator module.
\item \LightCTSLS{} that substitutes the last-shot compression with the classical full-shot aggregation method (see Figure~\ref{fig:ctsoverview}(a)), 
\item \LightCTSM{} that substitutes the L-MHA component of the GL-Former with the standard MHA component.
\item \LightCTSF{} that substitutes the L-FFN component of the GL-Former with the standard FFN component.
\item \LightCTSL{} that substitutes all local attention blocks with global attention blocks.

\end{itemize}
We also introduce three other \LightCTS{} variants to assess our design choices:
\begin{itemize}[leftmargin=*]
    \item \LightCTSA{} that adopts the alternate stacking pattern instead of the proposed plain stacking.
    \item \LightCTSCGCN{} that substitutes the GL-Former with Chebyshev GCNs~\cite{defferrard2016convolutional} as the S-operator module.
    \item \LightCTSDGCN{} that substitutes the GL-Former with Diffusion GCNs~\cite{gasteiger_diffusion_2019} as the S-operator module.
\end{itemize}

Table~\ref{tab:ablation} shows the results. We make the following observations.
1) \LightCTS{} achieves almost the best accuracy with much fewer FLOPs and parameters when all the lightweight techniques are deployed, implying that the proposed modules (i.e., L-TCN and GL-Former) and the last-shot compression are more efficient than their standard counterparts.
2) \LightCTSL{} is inferior to \LightCTS{} in terms of accuracy, indicating that the local attention block utilizing prior knowledge of spatial information does help achieve better forecasting performance.
3) The comparison between \LightCTS{} and \LightCTSA{} indicates that the plain stacking strategy is much better at reducing overheads than the standard alternate stacking strategy, while simultaneously improving the forecasting accuracy.
4) When comparing \LightCTSCGCN{}, \LightCTSDGCN{}, and \LightCTS{}, we observe that although \LightCTS{} consumes more resources, its accuracy surpasses the GCN-based models' by a large margin. Thus, we choose the Transformer-based S-operators in \LightCTS{}.

\subsection{Studies on Memory Constraints}
\label{ssec:resource}
We evaluate the performance of \LightCTS{} under memory constraints of 3Mb, 2.5Mb, and 2Mb, as commonly found in commodity MCUs ~\cite{STMCU}. 
We include representative baselines: \AutoCTS{} (the most accurate), \AutoCTSKDF{}/\AutoCTSKDP{} (KD variants of \AutoCTS{}), \AGCRN{} (the least peak memory use), and \GWNet{} (the lowest latency). 
For fair comparisons, we adjust the embedding size $\mathtt{D}$ for all models to fit into the constrained memory while maintaining their structures and components. Results on PEMS08 are presented in Table ~\ref{tab:memory}. Results on other datasets are available elsewhere~\cite{lightcts}. 

Even with the lowest possible embedding size, \AutoCTS{} and its variants are unable to comply with the memory constraints due to their large intermediate results. This is concrete evidence of the inapplicability of these models to resource-constrained devices such as MCUs.
While \GWNet{} and \AGCRN{} are able to meet the memory constraints, they experience significant accuracy loss. In contrast, \LightCTS{} shows the least accuracy degradation and surpasses the baseline models significantly with the lowest latency under all studied constraints.
These results demonstrate the need for specialized lightweight designs as state-of-the-art CTS models without such considerations fail to achieve satisfactory accuracy when downscaled for low-memory settings.

\begin{table}[ht]
\caption{Models vs memory constraints on PEMS08.}
\resizebox{\linewidth}{!}{

\begin{tabular}{cc||cccc||ccc}
\toprule
\begin{tabular}[c]{@{}c@{}}Mem Constraint\\ (unit: Mb) \end{tabular} & Model & \begin{tabular}[c]{@{}c@{}}FLOPs\\ (unit: M)\end{tabular} & \begin{tabular}[c]{@{}c@{}}Params\\ (unit: K)\end{tabular}& \begin{tabular}[c]{@{}c@{}}Latency\\ (unit: s)\end{tabular}& \begin{tabular}[c]{@{}c@{}}Peak Mem\\ (unit: Mb)\end{tabular}& MAE & RMSE & MAPE  \\ \midrule
& \LightCTS{} & \textbf{70} & 177  & \textbf{0.4} & 2.8 & \textbf{14.63} & \textbf{23.49} & \textbf{9.43\%}  \\ 
3& \GWNet{} &328 & 178  & 1.0 & 2.9 & 17.40 & 27.34 & 11.14\%  \\ 
& \AGCRN{}    & 726 & \textbf{150}  & 3.2 & \textbf{2.6}& 15.95 & 25.22 & 10.09\%  \\ \midrule

& \LightCTS{}  &  \textbf{42} & 103  & \textbf{0.3} & 2.5 &  \textbf{14.82} & \textbf{23.78} & \textbf{9.66\%}  \\ 
2.5& \GWNet{}& 137 & \textbf{49}  & 0.8 & \textbf{2.4}& 18.00 & 28.20 & 11.49\%  \\ 
 & \AGCRN{}    & 586 & 115 & 2.9 & 2.5 & 16.72 & 26.26 & 11.27\%  \\ \midrule 

& \LightCTS{}  & \textbf{8} & 15  & \textbf{0.2} & 2.0  &  \textbf{16.70} & \textbf{26.28} & \textbf{10.75\%} \\  
2 & \GWNet{} & 40 & \textbf{10}  & 0.6  & 2.0& 19.38 & 30.14 & 13.58\% \\  
  & \AGCRN{}    & 100 & 11 & 2.2 & 2.0 & 19.17 & 29.71 & 13.06\%  \\ 

   \bottomrule

\end{tabular}
}
\label{tab:memory}
\end{table}

%% file: content/6.Related_Work.tex
\section{Related Work}
\label{sec:related}

\noindent\textbf{DL-based Models for CTS Forecasting}.
Deep learning models dominate CTS forecasting. Different studies involve different S/T-operators. GCNs and GCN variants are the most common S-operators~\cite{li2018diffusion,yu2018spatio,bai2020adaptive,wu2020connecting,chen2022multi}. In addition to building graphs using prior knowledge in GCNs, learned graphs~\cite{wu2019graph,wu2020connecting,chen2022multi} and adaptive graphs~\cite{bai2020adaptive} demonstrate advantages in capturing dynamic and implicit correlations among time series. 
Further, TCNs~\cite{yu2018spatio,wu2019graph,wu2020connecting,guo2019attention,huang2020lsgcn,chen2022multi} and RNNs~\cite{bai2020adaptive,chen2020multi,chang2018memory,li2018diffusion,shih2019temporal} are the most widely adopted T-operators. 
Recently, the Transformer and its variants have been employed as S-operators~\cite{xu2020spatial,park2020st,zhou2021informer,wu2021autocts} and T-operators~\cite{xu2020spatial,park2020st,zhou2021informer,wu2021autocts}, due to the powerful correlation modeling abilities of Transformers. Neural architecture search (NAS) has been introduced to automatically select appropriate S/T-operators, resulting in competitive performance without having to design a CTS forecasting model manually \cite{wu2021autocts,pan2021autostg}.

All existing studies focus on improving forecasting accuracy. However, progress is slowing down and becoming marginal (see Tables~\ref{tab:multiresult} and~\ref{tab:singleresult}). In contrast, \LightCTS{} contributes lightweight S/T-operators and enables lightweight CTS models (w.r.t. computation and model size) without compromising forecasting accuracy. In this sense, \LightCTS{} renders forecasting more cost-effective and extends its potential applicability to edge devices in CPSs.

\noindent\textbf{Lightweight DL Models}.
Developing light DL models is motivated by the requirements of real-time and mobile applications. There are two streams of related work \cite{chen2020deep}: compressing well-trained big models and designing lightweight models from scratch. 
The first stream has been well studied in areas like  CV~\cite{krizhevsky2012imagenet,han2015learning,yu2018nisp} and NLP~\cite{michel2019sixteen,jiao2019tinybert,wang2020minilm} fields. 
However, \LightCTS{} falls outside this stream as there are no compelling well-trained CTS models.

Next, impressive advances in the design of lightweight models from scratch have also been widely reported in CV~\cite{zhang2018shufflenet,DBLP:conf/cvpr/SandlerHZZC18,tan2019efficientnet,howard2017mobilenets,howard2019searching,wu2018shift}. A popular operator is the depth-wise separable convolution that decouples the standard convolution into intra- and inter-feature map computation. It serves as the basic block of the famous MobileNets~\cite{howard2017mobilenets} and Xception \cite{chollet2017xception}. 
A follow-up work is the inverted bottleneck structure~\cite{DBLP:conf/cvpr/SandlerHZZC18} that finds a narrow-wide-narrow convolution to achieve reduced computations while competitive accuracy. 
From a different perspective, EfficientNet~\cite{tan2019efficientnet} aims to scale existing modules to meet certain constraints rather than designing new efficient modules.
However, the aforementioned lightweight modules cannot be applied directly to CTS forecasting because of the inherently different data structures and tasks involved. 
For example, these methods focus mainly on simplifying 2D and 3D convolutions for extracting local features of image and video data, while CTS models require uncovering long-term temporal dynamics and non-uniform spatial correlations. Given this gap, we identify potential directions to achieve lightness of CTS forecasting based on a careful study of existing CTS models.
\LightCTS{} offers a plain stacking architecture together with a last-shot compression to efficiently deal with the heterogeneity of S/T-operators, which is unique compared to models used in CV. 
We also design L-TCN and GL-Former according to the exclusive characteristics of temporal dynamics and spatial correlations in CTS.

%% file: content/7.Conclusion_and_Future_Work.tex
\section{Conclusion}
\label{sec:conclusion}

We present \LightCTS{}, a new framework for lightweight forecasting of correlated time series (CTS) that achieves comparable accuracy to state-of-the-art CTS forecasting models but consumes much fewer computational resources.
\LightCTS{} integrates a set of novel computational cost reduction techniques, notably a plain stacking architecture, the L-TCN (Light Temporal Convolutional Network) and GL-Former (GlobalLocal TransFormer) modules for extracting spatio-temporal features, and a last-shot compression scheme for reducing redundant, intermediate features.
Comprehensive experiments offer evidence that \LightCTS{} is capable of providing state-of-the-art CTS forecasting accuracy with much fewer FLOPs and parameters than existing CTS forecasting proposals.
